\documentclass[nonacm]{acmart}
\usepackage{wrapfig}
\usepackage{geometry}
\usepackage{paralist}
\usepackage{caption}
\usepackage{subcaption}
\usepackage{enumitem}
\usepackage{comment}
\usepackage{tabularx}
\usepackage{soul}
\usepackage[usestackEOL]{stackengine}
\usepackage{array,ragged2e}

\newsavebox{\measurebox}

\AtBeginDocument{%
  }

\begin{document}

\strutlongstacks{T}

\title{On the Robustness of Explanations of Deep Neural Network Models: A Survey}
\author{Amlan Jyoti}
\authornote{These authors contributed equally to this research, author names arranged in alphabetical order by first name.}
\email{cs20mds14007@iith.ac.in}
\author{Karthik Balaji Ganesh}
\authornotemark[1]
\email{cs19btech11001@iith.ac.in}
\author{Manoj Gayala}
\authornotemark[1]
\email{cs19btech11011@iith.ac.in}
\author{Nandita Lakshmi Tunuguntla}
\authornotemark[1]
\email{cs19btech11051@iith.ac.in}
\author{Sandesh Kamath}
\authornotemark[1]
\email{sandesh.kamath@cse.iith.ac.in}
\author{Vineeth N Balasubramanian}
\email{vineethnb@cse.iith.ac.in}
\affiliation{%
  \institution{Indian Institute of Technology, Hyderabad}
  \streetaddress{Kandi}
  \city{Sangareddy}
  \state{Telangana}
  \country{India}
  \postcode{502284}
}








\renewcommand{\shortauthors}{Jyoti et al.}

\begin{abstract}
Explainability has been widely stated as a cornerstone of the responsible and trustworthy use of machine learning models. With the ubiquitous use of Deep Neural Network (DNN) models expanding to risk-sensitive and safety-critical domains, many methods have been proposed to explain the decisions of these models. Recent years have also seen concerted efforts that have shown how such explanations can be distorted (attacked) by minor input perturbations. While there have been many surveys that review explainability methods themselves, there has been no effort hitherto to assimilate the different methods and metrics proposed to study the robustness of explanations of DNN models. In this work, we present a comprehensive survey of methods that study, understand, attack, and defend explanations of DNN models. We also present a detailed review of different metrics used to evaluate explanation methods, as well as describe attributional attack and defense methods. We conclude with lessons and take-aways for the community towards ensuring robust explanations of DNN model predictions.
\end{abstract}

\maketitle

\section{Introduction}
\label{sec:intro}
Deep neural network (DNN) models have enjoyed tremendous success across application domains within the broader umbrella of artificial intelligence (AI) technologies. However, their ``black-box'' nature, coupled with their extensive use across application sectors —including safety-critical and risk-sensitive ones such as healthcare, finance, aerospace, law enforcement, and governance—has elicited an increasing need for explainability, interpretability, and transparency of decision-making in these models \cite{xaitutorial,molnar2019,samek_explainable_2019,TjoaG21}. With the recent progression of legal and policy frameworks that mandate explaining decisions made by AI-driven systems (for example, the European Union’s GDPR Article 15(1)(h) and the Algorithmic Accountability Act of 2019 in the U.S.), explainability has become a cornerstone of responsible AI use and deployment.

Existing efforts on explaining the predictions of machine learning models can be broadly categorized as: local and global methods, model-agnostic and model-specific methods, causal and non-causal methods, or as post-hoc and ante-hoc (intrinsically interpretable) methods \cite{molnar2019,xaitutorial}. While many methods have been proposed for explainability and also captured through many literature surveys  \cite{TjoaG21,islam2021explainable,burkart2021survey,DosilovicBH18,CapuanoFLS22,Rojat22,Das20,SahakyanAR21,AdadiB18,carvalho2019machine,AliciogluS22}, there has been very little effort on presenting a review of techniques for the evaluation of explanation methods. The evaluation of explainability of DNN models is known to be a challenging task, necessitating such an effort. From another perspective, while there have been many surveys of literature on adversarial attacks and robustness \cite{LongGXZ22,AlsmadiANASVKAA22,Silva20,Meng20,Goyal22,SunTZ18,Akhtar18,Li18,Sun18,Wang19,ZhouHLHG19,WangLKTL19,Serban20,Vakhshiteh20,Huq20,Chaubey20,ZhangBLKWK21,WangWL21,KongXWHNL21,ChakrabortyADCM21,AkhtarMKS21,Wang22,MichelJE22,KavianiHS22,DingX20} -- which focus on attacking the predictive outcome of these models, there have been no effort so far to study and consolidate existing efforts on attacks on explainability of DNN models. Many recent efforts have demonstrated the vulnerability of explanations (or attributions\footnote{We use the terms explanations and attributions interchangeably in this work.}) to human-imperceptible input perturbations across image, text and tabular data \cite{ghorbani2017fragile,slack2019posthoc,ivankay2022text,sinha2021NLP,lakkaraju2020fool,zhang2018fire,dombrowski2019geometry}. Similarly, there have also been many efforts in recent years in securing the stability of such explanations in \cite{HuaiLMYZ22,dombrowski2019geometry,SinghKMSBK20,alvarez2018robustness,WangWRMFD20,SchwartzAK20,DombrowskiAMK22,ChalasaniC00J20,Chen0RLJ19,ivankay2020far,SarkarSB21,ManglaSB20}. These efforts have however remained disparate depending on the domain and type of data studied. The growing importance for robust explanations of DNN models in practice requires a consolidation of these efforts, thus enabling the community of researchers to build on existing efforts in a more holistic manner. In this work, we seek to address this impending need through a survey of approaches that study, understand, attack and defend explanations of DNN models.

\begin{figure}[t]
\centering
\begin{subfigure}{.17\textwidth}
  \includegraphics[width=1.0\textwidth, height=0.15\textheight]{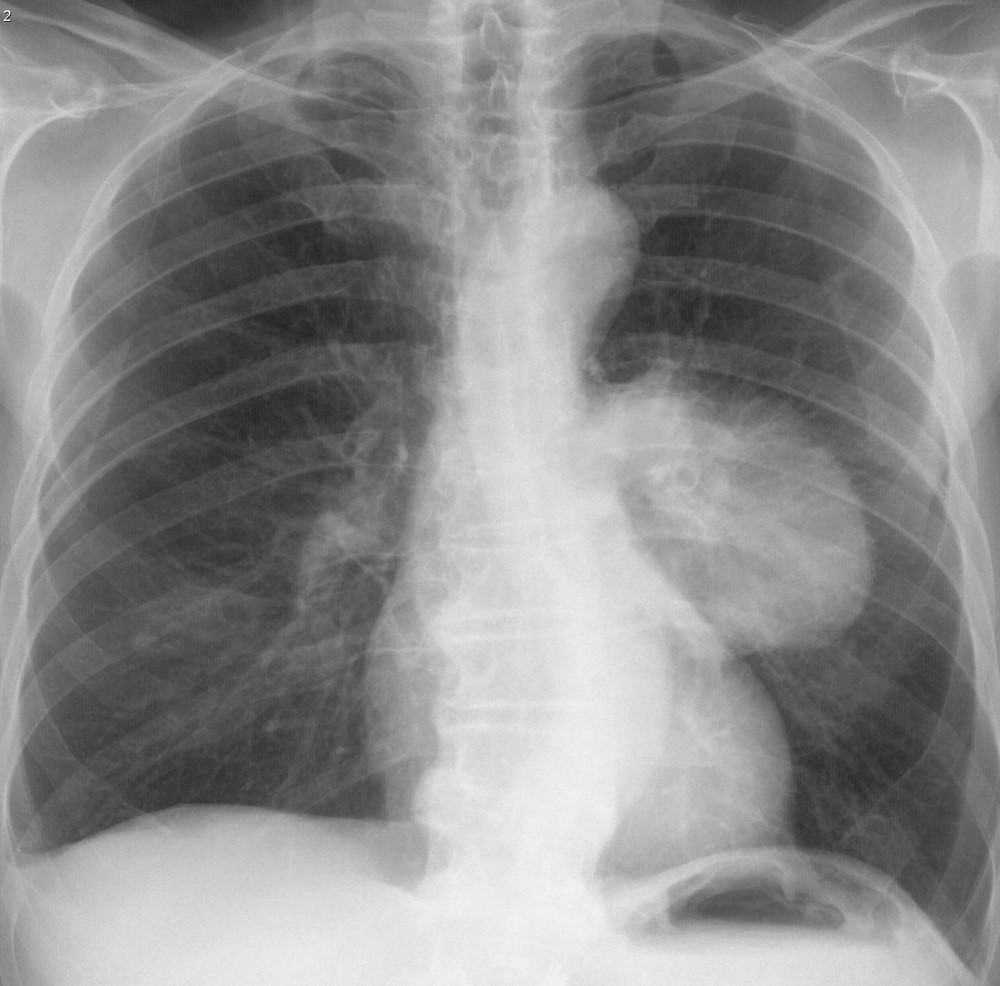}  
  \label{fig:1sub-first}
\end{subfigure}
\begin{subfigure}{.17\textwidth}
  \includegraphics[width=1.0\textwidth, height=0.15\textheight]{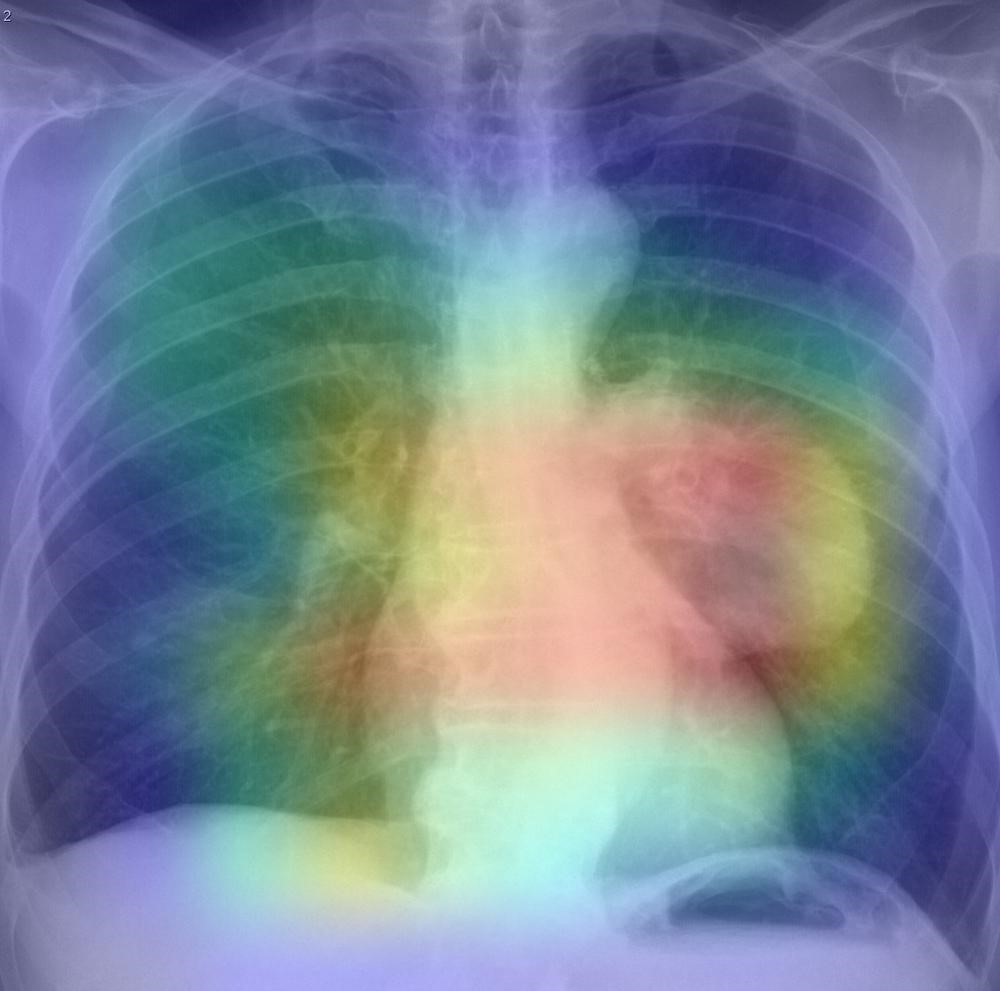}  
  \label{fig:1sub-third}
\end{subfigure}
\begin{subfigure}{.17\textwidth}  
  \includegraphics[width=1.0\textwidth, height=0.15\textheight]{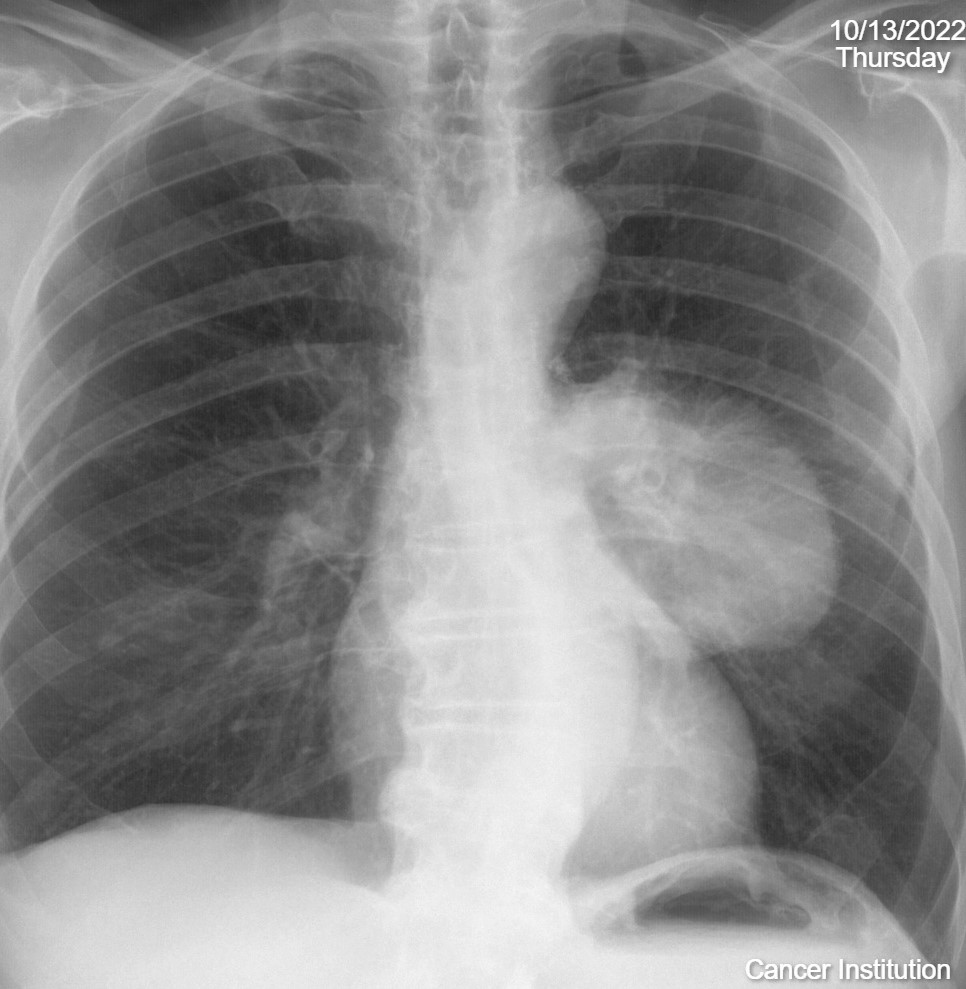}  
  \label{fig:1sub-second}
\end{subfigure}
\begin{subfigure}{.17\textwidth}
  \includegraphics[width=1.0\textwidth, height=0.15\textheight]{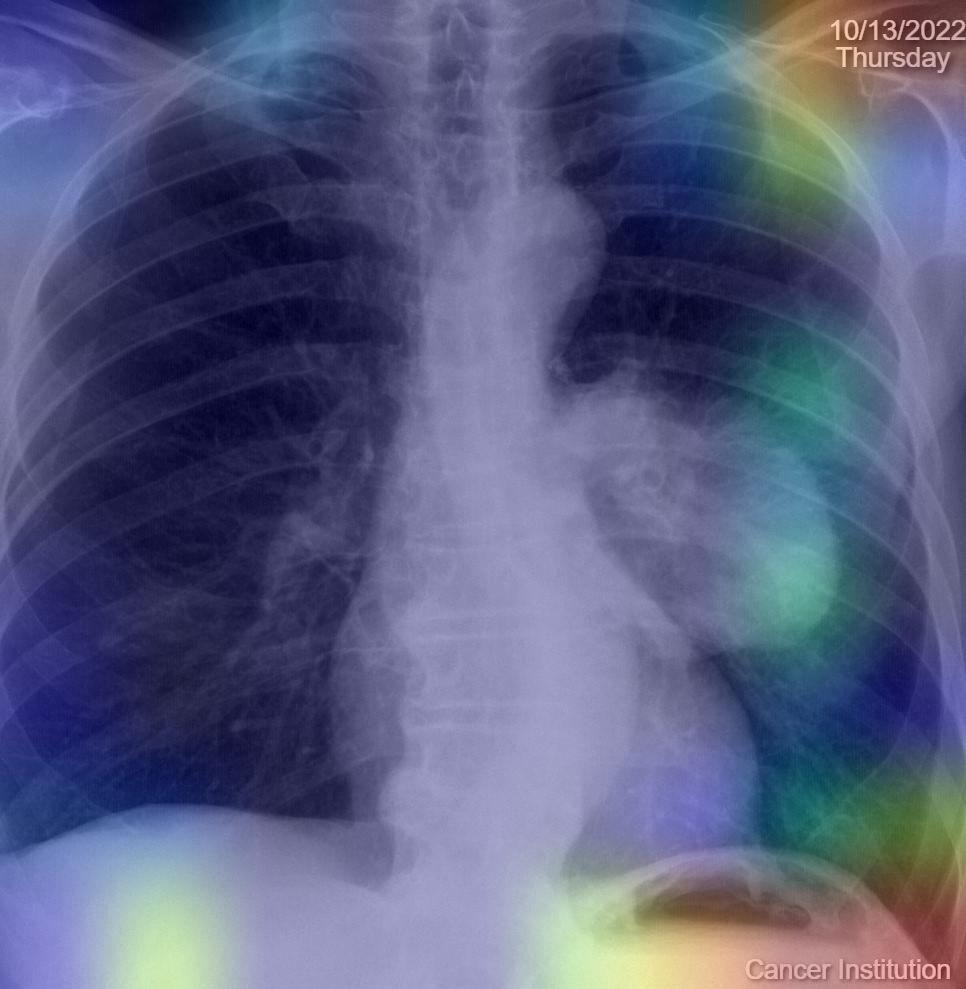}  
  \label{fig:1sub-fourth}
\end{subfigure}\\
\begin{subfigure}{.17\textwidth}
  \includegraphics[width=1.0\textwidth, height=0.15\textheight]{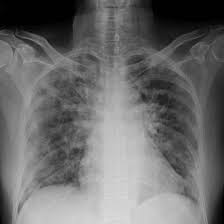}  
  \label{fig:2sub-first}
\end{subfigure}
\begin{subfigure}{.17\textwidth}
  \includegraphics[width=1.0\textwidth, height=0.15\textheight]{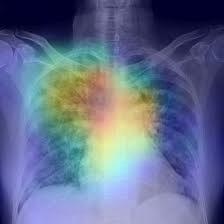}  
  \label{fig:2sub-third}
\end{subfigure}
\begin{subfigure}{.17\textwidth}  
  \includegraphics[width=1.0\textwidth, height=0.15\textheight]{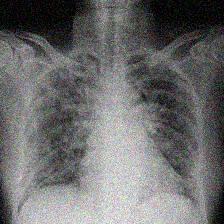}  
  \label{fig:2sub-second}
\end{subfigure}
\begin{subfigure}{.17\textwidth}
  \includegraphics[width=1.0\textwidth, height=0.15\textheight]{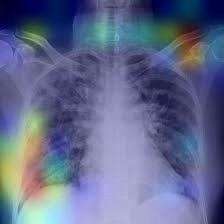}  
  \label{fig:2sub-fourth}
\end{subfigure}
\vspace{-14pt}
\caption{\textit{(Top row,a-d from Left to Right)} (a) Original x-ray of a lung cancer patient; (b) Attribution map of original x-ray highlighting relevant areas of the x-ray; (c) Watermarks added to x-ray in (a); (d) Modified attribution map obtained of x-ray with watermarks with prediction remaining the same. The model accurately predicts the presence of lung cancer in (a) and (c), but the corresponding attribution maps in (b) and (d) respectively are not robust. Note how the areas with the watermark are highlighted here, indicating that the model uses the watermarks to make the decision, making the model unreliable. \textit{(Bottom row,a-d from Left to Right)} (a) Original x-ray of a COVID-19 infected patient; (b) Attribution map of original x-ray highlighting relevant areas of the x-ray; (c) Human-imperceptible perturbation added to x-ray in (a); (d) Modified attribution map of perturbed x-ray in (c). The model accurately predicts the presence of COVID-19 infection in (a) and (c), but the corresponding attribution maps in (b) and (d) respectively are not robust. If a physician relied on such modified attribution maps for understanding the disease, it could be life-threatening for the patient.}
\vspace{-8pt}
\label{fig_intro}
\end{figure}

Beyond being generally useful, the stability and robustness of explanations of DNN model predictions become extremely essential in risk-sensitive and safety-critical application domains. Figure \ref{fig_intro} presents an illustration of a practical example in a healthcare setting. As DNN models are more widely deployed, the attributions (or saliency maps, when the input is an image) may become common vehicles of reasoning, which are used by end-users to make decisions. In the presented example, if a physician uses the attribution map to make a final diagnosis, relying on watermarks or distorted attribution maps may be life-threatening for a patient. Robust explanations are similarly essential in many other application domains including autonomous navigation, security, governance, and finance.

Our key contributions in this work are as follows: (i) We present a comprehensive survey of methods that study, understand, attack and defend explanations of DNN models. To the best of our knowledge, this is the first such effort on this topic; (ii) We present an overarching review of different approaches for evaluation of explanation methods (Section \ref{sec:eval-explain}) including widely used properties and axioms, and how they are typically measured; (iii) We summarize the different evaluation metrics that have been proposed hitherto for the robustness of explanations in DNN models (Section \ref{sec:evaluate-robustness-explanation}); (iv) We present detailed descriptions of different attributional attack and defense methods (Sections \ref{sec:attack-explain} and \ref{sec:ensure-robust}, respectively); (v) Considering the vast body of literature that is present for adversarial robustness, we present connections between adversarial training and robust explanations (Section \ref{sec:adv-connect}); and (vi) We finally present lessons and take-aways towards robust explanations for DNN models from our in-depth study of existing literature (Section \ref{sec:conclusion}), in order to benefit readers, researchers and practitioners in the field.

\vspace{-5pt}\section{Overview of Explainability Methods} \label{sec:overview-exp-methods}
We begin our exposition with a brief overview of explainability methods.
Numerous methods have been proposed over the last few years to explain the importance of an input feature to the prediction of a DNN model. These methods include simple visualization of weights and neurons \cite{SimonyanVZ13,erhan2009visualizing}, deconvolutional networks \cite{zeiler2014visualizing}, guided backpropagation \cite{SpringenbergDBR14}, Layer-wise Relevance Propagation (LRP) \cite{BinderMLMS16}, CAM (Class Activation Mapping) \cite{ZhouKLOT16}, GradCAM (Gradient-weighted Class Activation Mapping) \cite{SelvarajuCDVPB17}, GradCAM++ \cite{ChattopadhyaySH18}, LIME \cite {attr2016perturb_3_lime,PengM21a}, QII (Quantitative Input Influence) \cite{DattaSZ16}, DeepLIFT \cite{ShrikumarGK17}, Integrated Gradients (IG) \cite{SundararajanTY17}, Layered Integrated Gradients \cite{MudrakartaTSD18}, SHAP \cite{StrumbeljK14}, GradientSHAP \cite{LundbergL17}, Noise Tunnel \cite{AdebayoGMGHK18}, SmoothGrad \cite{SmilkovTKVW17}, Conductance \cite{DhamdhereSY19}, Testing with Concept Activation Vectors (TCAV) \cite{kim2018interpretability}, Average Causal Effect (ACE) \cite{chattopadhyay2019neural}, etc. Dedicated software packages that implement such methods are also available commonly, with the popular ones including Pytorch-Captum\footnote{https://github.com/pytorch/captum} and Tensorflow-Responsible AI\footnote{https://www.tensorflow.org/responsible\_ai} libraries. 

These existing methods can be categorized in many ways. Based on the input used, existing libraries\footnote{https://captum.ai/docs/attribution\_algorithms} categorize methods as: \textit{primary} attribution methods (use only input) \cite{SundararajanTY17,LundbergL17,ShrikumarGK17,SpringenbergDBR14,zeiler2014visualizing,StrumbeljK14,PengM21a}, \textit{layer} attribution methods (use layerwise weights) \cite{SelvarajuCDVPB17,DhamdhereSY19,LeinoSDFL18}, or \textit{neuron} attribution methods (based on a single neuron) \cite{DhamdhereSY19}. Adding noise to aid the process is termed as the Noise Tunnel method \cite{SmilkovTKVW17,AdebayoGGK18}. Based on how the attributions are obtained, they are sometimes also categorized as \textit{gradient-based} \cite{SimonyanVZ13,erhan2009visualizing,zeiler2014visualizing,SpringenbergDBR14,BinderMLMS16,SelvarajuCDVPB17,ChattopadhyaySH18,SundararajanTY17,MudrakartaTSD18} and \textit{perturbation-based} \cite{StrumbeljK14,LundbergL17,PengM21a}\footnote{https://captum.ai/docs/algorithms\_comparison\_matrix}. They can also be categorized as \textit{local} and \textit{global} methods (depending on whether they apply to a single data sample or a dataset as a whole), \textit{model-agnostic} and \textit{model-specific} methods (depending on whether the method is specific to a model or general), \textit{causal} and \textit{non-causal} methods (depending on whether causal perspectives are considered), or as \textit{post-hoc} and \textit{ante-hoc} (intrinsically interpretable) methods \cite{molnar2019,xaitutorial}. 

For more details, we request the interested reader to refer to \cite{DosilovicBH18,AliciogluS22,burkart2021survey,islam2021explainable,Das20,AdadiB18} for general surveys on explainability methods, or for specific domains refer to: medical \cite{TjoaG21}, embedded systems \cite{vidot2021certification}, multimodal \cite{JoshiWK21}, time series \cite{Rojat22}, cybersecurity \cite{CapuanoFLS22}, and tabular data \cite{SahakyanAR21}.

\vspace{-5pt}\section{Evaluation of Explainability Methods} \label{sec:eval-explain}

    




    

In order to understand and analyze explanation methods, it is essential to be able to evaluate them. This evaluation is challenging for a subjective concept like explainability since different humans might interpret the same information differently. Further, ground truth data often is hard to obtain for explanations in real-life scenarios. Consequently, various metrics to evaluate explanations have been developed over the years, serving different contexts and purposes. This section shall look at these different existing ways to evaluate explanations.

Section \ref{subsec:notations} provides a list of notations that serve as preliminaries for this survey. Section \ref{subsec:properties-and-axioms} looks at the various qualitative properties and axioms desired in a good explanation, while Section \ref{subsec:evaluate-quality-explanation} focuses on the different quantitative metrics proposed over the years to evaluate the quality of an explanation method.

\vspace{-6pt}\subsection{Notations}\label{subsec:notations}

Table \ref{table:notations} serves as an index for the notations used during this survey.

\begin{table}[h!]
\centering
\resizebox{\textwidth}{!}{
\begin{tabular}{||c | l||} 
 \hline
 Notation & {\centering Physical Significance} \\ [0.5ex] 
 \hline\hline
 $\mathcal{D}$ & Data distribution from which sample $(x,y)$ is drawn\\
 $x$ & Sample input $x \in \mathcal{D}$ \\
 $y$ & Label corresponding to sample $x$ (supervised learning)\\
 $x'$ & Represents the perturbed input or $x+\delta$; generally in the neighborhood of $x$ ($\mathcal{N}_x$) with radius $\delta$ \\
 $\mathcal{F}$ & The model function, $\mathcal{F} : x \to y$; $\mathcal{F}(x)$ is the model prediction for sample input $x$ \\
 $\mathcal{I}$ & The explanation functional \\
 $\mathcal{L}$ & The loss function used to evaluate the model, in general cross-entropy \\ [1ex] 
 \hline
\end{tabular}
}
\caption{Index of notations used in this survey}
\label{table:notations}
\vspace{-9mm}
\end{table}

\vspace{-6pt}\subsection{Properties and Axioms}\label{subsec:properties-and-axioms} 

Evaluating explanations has proven challenging, considering the diverse range of methods employed. The first step in this endeavor is defining properties desired in a \textit{good} explanation. Many different properties have been proposed over the years, and some of these include:

\begin{itemize}[leftmargin=*]
    
    \item \textbf{Sparsity} \cite{doshi2017towards, ajalloeian2021evaluation, lisboa2021coming}  : A good explanation identifies sparse components of the original input that have the greatest influence on model predictions, i.e., few of the most relevant features are assigned maximum value. Sparse explanations are more concise and make the explanations more comprehensible to the end user.
    
    \item \textbf{Fidelity} \cite{carvalho2019machine, robnik2018perturbation, yeh2019fidelity, fel2022good, hsieh2020evaluations}: Fidelity corresponds to how well an explanation reflects the behavior of the underlying model; in other words, it indicates how faithful the explanation is to the model. Fidelity is based on the simple concept that features with the highest attribution value are most relevant and necessary, i.e., perturbing or removing these features causes a considerable difference in the model's prediction score.

    \item \textbf{Stability / Sensitivity} \cite{lisboa2021coming, bhatt2020evaluating, alvarez2018towards, carvalho2019machine, robnik2018perturbation} : The notion of stability focuses on the consistency of explanations for similar inputs, i.e., explanations must be robust to local perturbations of the input. Stable explanations do not vary too much between similar input samples unless the model's prediction changes drastically \cite{carvalho2019machine}. The term \textit{sensitivity} is often used interchangeably to refer to stability \cite{bhatt2020evaluating}, and low sensitivity corresponds to greater stability. 
    
    \item \textbf{Comprehensibility} \cite{carvalho2019machine, robnik2018perturbation}: Comprehensibility represents how well humans understand the explanations. It is one of the most important properties but also one of the most difficult to define and measure.
    
    \item \textbf{Algorithmic Complexity} \cite{carvalho2019machine, robnik2018perturbation}: Generating explanations in a computationally efficient or inexpensive manner is vital; a good explanation that requires a lot of time and resources for computation is not useful from a practical perspective.

    \item \textbf{Efficiency} \cite{carvalho2019machine, ruping2006learning}: Efficiency corresponds to the time needed for a user to understand the explanation; the more efficient an explanation is, the faster a user can grasp it.
    
    \item \textbf{Repeatability \& Reproducibility} \cite{arun2020assessing}: A good explanation method is expected to be repeatable and reproducible. This can be evaluated by comparing explanations from different randomly initialized instances of the same model trained until convergence (intra-architecture) and by comparing explanations from different model architectures trained until convergence on the same task (inter-architecture). A good explanation method generates fairly robust explanations across all these different instances and architectures as long as the model is trained on the same data for the same task.
    
    \item \textbf{Generalizability \& Consistency} \cite{fel2022good}: As defined by \citet{fel2022good}, \textit{generalizability} is a measure of how generalizable an explanation is and how well it reflects the underlying mechanism by which a predictor makes a decision. \textit{Consistency}, on the other hand, is the extent to which different predictors trained on the same task do not exhibit logical contradictions, i.e., the degree to which different predictors are consistent with each other. Intuitively, the notion of generalizability indicates that for different images correctly classified under the same class, a trustworthy explanation must point to the same evidence across all images. Conversely, consistency holds when, for misclassified images, the explanation must point to evidence that is different from the ones used to classify the images correctly.
\end{itemize}
\vspace{-3pt}


As we can observe, quite a few properties have been proposed over the years that are desired in good explanation methods. For a highly subjective field such as explainability, it is desirable to have an objective basis for evaluating the explanation methods. Axioms are one such example; generally associated with a mathematical formulation, they can serve as an objective indicator. Explanation methods can be evaluated based on whether or not they satisfy a given axiom, which in turn provides information about the properties of the explanation method itself. Some of the well-known axioms proposed over the years are:

\vspace{-3pt}
\begin{itemize}[leftmargin=*]
    \item \textbf{Implementation Invariance} \cite{SundararajanTY17}: Functional equivalence describes two networks whose outputs are equal for all inputs, regardless of their implementation. Implementation Invariance is an axiom proposed for explanation methods, requiring attributions to be identical for two functionally equivalent networks.
    
    \item \textbf{Sensitivity} \cite{SundararajanTY17}: \citet{SundararajanTY17} define that an attribution method satisfies the Sensitivity axiom when for every input and baseline that differ in only a single feature but have different predictions, the attribution corresponding to that distinguishing feature is non-zero. Conversely, if the function implemented by a deep neural network does not depend on some variable, then zero attribution must be assigned to that variable to indicate the same.
    
    \item \textbf{Input Invariance} \cite{kindermans2017}: Extending the principles of implementation invariance, \citet{kindermans2017} posit that given a DNN, a second DNN can be constructed to handle a constant shift in input. In other words, it produces the same output for the constant-shifted input as the first network did on the initial input. Input Invariance proposes that the saliency maps must mirror the model's sensitivity to this shift in input. The attribution maps from different methods are compared for the two networks. It is discovered that for methods based on reference points, both the reference point and the shift in input influence the final attribution map.
    
    \item \textbf{Completeness} \cite{SundararajanTY17, ancona2017towards, yeh2019fidelity, lisboa2021coming}  : An explanation satisfies the Completeness axiom when the sum of the attributions is equal to the difference in the model's output at a given input and the baseline input i.e. given a model $\mathcal{F}$, input $x$ and baseline $x_0$ :  $\space\Sigma Attr(x) = \mathcal{F}(x) - \mathcal{F}(x_0)$. Completeness can thus be interpreted as a more general form of the Sensitivity axiom.
    
    \item \textbf{Honegger's Axioms} \cite{honegger2018shedding} : \citet{honegger2018shedding} proposed three axioms that relate objects with their corresponding explanations. These are: \textit{Identity} (identical objects must have identical explanations), \textit{Separability} (non-identical objects cannot have identical explanations) and \textit{Stability} (similar objects must have similar explanations).

\end{itemize}

It is important to remember that our primary motivation is to evaluate explanation methods based on the properties they satisfy to determine the \textit{goodness} of an explanation. Table \ref{table:axioms-and-properties} provides a tabular view of this association, relating the axioms to the properties they encapsulate.


\subsection{Evaluating the Quality of Explanation methods}\label{subsec:evaluate-quality-explanation}

In Section \ref{subsec:properties-and-axioms}, we looked at different properties and axioms proposed for explanation methods over the years. However, axioms lead to a black-or-white (or 0-or-1) interpretation where an explanation method either satisfies an axiom or does not. In practice, we might require an option in-between (grey). This leads us towards objective quantitative metrics, which have a two-fold advantage. First, they generate a numerical value that allows a relative comparison between different explanation methods. Second, a quantitative value allows us to understand the extent to which a method satisfies a given metric or property and is no longer a simplistic black-or-white answer. \\
\begin{wraptable}{r}{0.6\textwidth}
\vspace{2pt}
\resizebox{0.6\textwidth}{!}{
\begin{tabular}{||c | c||} 
 \hline
 Axiom & Properties \\ [0.5ex] 
 \hline\hline
 Implementation Invariance & Repeatability and Reproducibility \\ 
 Sensitivity & Fidelity \\
 Input Invariance & Generalizability, Repeatability and Reproducibility \\
 Completeness & Fidelity \\
 Hoenegger's Identity & Generalizability \\
 Hoenegger's Separability & Consistency \\
 Hoenegger's Stability & Stability \\ [1ex] 
 \hline
\end{tabular}}
\caption{Associating the axioms to properties enlisted in Section \ref{subsec:properties-and-axioms}}
\vspace{-20pt}
\label{table:axioms-and-properties}
\end{wraptable}
Since the domain of explainability is subjective, the search for objective evaluation metrics is still an open field of research. While there is no single metric yet to evaluate explanations across all benchmarks, different metrics have been introduced to measure different properties over the years. In this section, we provide a general overview of these metrics. Table \ref{table:metrics-math-definitions} enlists the mathematical formulations for each of these metrics. The quantitative metrics are:

\vspace{-4pt}
\begin{itemize}[leftmargin=*]
    
    
    

    
    
    
    \item \textbf{Gini Index} \cite{ajalloeian2021evaluation}: The Gini Index is a popular metric used to measure the sparsity of a non-negative vector \cite{hurley2009comparing}. \citet{ajalloeian2021evaluation} apply the Gini Index on the absolute value of the flattened explanation map to evaluate the sparsity of the corresponding map. A larger Gini Index suggests a more concise (sparser) explanation; thus, this metric provides us with a direct estimate of how sparse an explanation is.
   

 
    
    \item \textbf{Sensitivity-n} \cite{ancona2017towards}: \citet{ancona2017towards} propose a new metric Sensitivity-n for evaluating explanation methods. An attribution method is said to satisfy Sensitivity-n when the sum of the attributions in any subset of features of cardinality $n$ is equal to the difference in the prediction $\mathcal{F}$ caused by removing this subset of features (i.e., setting them to zero). One can observe that when $n = N$, where $N$ is the total number of features, the definition of \textit{Completeness} is obtained. Thus, Sensitivity-n can be considered a generalization of the Completeness axiom.

    \item \textbf{Fidelity} \cite{yeh2019fidelity, ajalloeian2021evaluation, fel2022good, bhatt2020evaluating, samek2016evaluating, petsiuk2018rise, rieger2020irof}: Fidelity quantifies the ability of an explanation to capture and reflect the change in the prediction value in response to significant perturbations and input masking. In other words, it attempts to estimate the relevance of different features, which is generally measured with reference to a specified baseline. Fidelity is perhaps the most researched metric for evaluating explanations. One such metric by \citet{bhatt2020evaluating} computes the correlation between the attribution of a subset and the change in prediction scores on masking (removing) that subset. Other metrics that also essentially measure the correlation between the drop in the prediction score and the explanation scores of the corresponding masked input variables have been defined \cite{samek2016evaluating, petsiuk2018rise, rieger2020irof}. A more general metric, Explanation Infidelity, was defined by \citet{yeh2019fidelity} and is defined in terms of a random variable $\mathcal{J} \in \mathbb{R}^d$ with an associated probability measure $\mu_{\mathcal{J}}$, which represents meaningful perturbations of interest. $\mathcal{J}$ can thus represent different perturbations around $x$, such as difference to a specific baseline ($\mathcal{J} = x - x_0$), difference to a noisy baseline ($\mathcal{J} = x - (x_0 + \epsilon)$, where $\epsilon \sim \mathcal{N}(0, \sigma^2)$), difference to a subset of baseline, etc. The infidelity metric can thus be considered a measure of the distance between the explanation and the model prediction. A lower infidelity value indicates an explanation method with better fidelity.
     
    
    
    
    \item \textbf{Smallest Sufficient Region (SSR)} \cite{dabkowski2017real}: The Smallest Sufficient Region objective is related to the concept of fidelity; the goal here is to find the smallest region of the image that allows a confident classification (hence, also sufficient). An analogous objective is Smallest Destroying Region. \citet{dabkowski2017real} introduced a new saliency metric to evaluate the quality of saliency maps based on the SSR objective. A low value for the saliency metric is a characteristic of a good saliency map.

    \item\textbf{Mean Generalizability (MeGe) \& Relative Consistency (ReCo)} \cite{fel2022good}: To evaluate how well a predictor's explanation generalizes from seen to unseen data points, \citet{fel2022good} define the following preliminaries: Given a dataset $\mathcal{D}$, it is separated into $k$ disjoint folds $V = \{V_1, V_2, ..., V_k\}$. $\mathcal{A}$ is a deterministic learning algorithm that maps any number of data points onto a function $\mathcal{F}$ from $X$ to $Y$. For each fold $V_i$, the corresponding predictor $\mathcal{F}_i$ is defined as $\mathcal{F}_i = A(V \symbol{92} V_i)$. Now, given an explanation functional $\mathcal{I}$, $\mathcal{I}_x^i$ is defined as $\mathcal{I}_x^i = \mathcal{I}(\mathcal{F}_i, x)$, and the distance value $\delta_x^{(i, j)} = d(\mathcal{I}_x^i, \mathcal{I}_x^j)$ is also defined, where $d(.,.)$ is a distance metric over the explanations. With these quantities, the complementary sets $S^{=}  = \{\delta_x^{(i, j)}: \mathcal{F}_i(x) = y \land \mathcal{F}_j(x) = y\}$ and $S^{\neq}  = \{\delta_x^{(i, j)}: \mathcal{F}_i(x) = y \oplus \mathcal{F}_j(x) = y\}$ are defined, and $S$ is defined as the union of these two sets i.e. $S = S^{=} \cup S^\neq$. Finally, to evaluate \textit{ReCo}, we require the True Positive Rate and the True Negative Rate, defined as $TPR(\gamma) = \frac{|\{\delta \in S^{=}: \delta \leq \gamma\}|}{|\{\delta \in S: \delta \leq \gamma\}|}$ and $TNR(\gamma) = \frac{|\{\delta \in S^{\neq}: \delta > \gamma\}|}{|\{\delta \in S: \delta >    \gamma\}|}$ respectively, where $\gamma \in S$ is a fixed threshold.
    
    
    With these preliminaries, the metrics of interest are defined in Table \ref{table:metrics-math-definitions}. We can see that a higher $MeGe$ score corresponds to a lower distance between the explanations, i.e., the explanation performs well even when a sample point is removed from the training set. In other words, it generalizes well to unseen data points. Similarly, $ReCo$ looks for a distance value that separates $S^{=}$ and $S^{\neq}$; the clearer the separation, the more \textit{consistent} the explanations are. A $ReCo$ value of 1 indicates complete consistency of the predictors' explanations, and 0 indicates complete inconsistency.
    
    
    
    
    
    
    \item \textbf{Robustness-S} \cite{hsieh2020evaluations}: Given a relevant feature set $S_r$, $Robustness{-}S_r$ and $Robustness{-}\bar{S_r}$ can be used to evaluate an explanation. A set $S_r$ which has a larger number of relevant features would correspond to a lower value of $Robustness{-}S_r$ and a higher value of $Robustness{-}\bar{S_r}$, i.e., it is based on the principle that even small perturbations to the most relevant features will have a great impact. This can also be quantitatively measured by plotting an AUC curve of $Robustness{-}S_r$ against different values of $k$, where the top-$k$ features from the evaluated explanation method are chosen as $S_r$.
    
    
    
    
    
    
    
    \item \textbf{Explanation Sensitivity} \cite{yeh2019fidelity, alvarez2018robustness, agarwal2022rethinking}: Another evaluation metric defined by \citet{yeh2019fidelity} is Max-Sensitivity, which estimates the change in the explanation in response to a perturbation in the input. For an explanation that is locally Lipschitz continuous \cite{alvarez2018robustness}, the max-sensitivity is also bounded. From the definition in Table \ref{table:metrics-math-definitions}, we can see that we ideally prefer lower values of max-sensitivity. Other metrics proposed to evaluate sensitivity include \textit{local stability} introduced by \citet{alvarez2018robustness} and \textit{relative stability} proposed by \citet{agarwal2022rethinking}, which evaluates the stability of an explanation with respect to changes in the model input, intermediate model representations as well as output logits. This change is measured in terms of the percentage change in the explanation with respect to the corresponding perturbation, hence referred to as \textit{relative stability}. Using percentage changes instead of absolute values also allows comparison across multiple explanation methods that vary in range and magnitude.
    
\end{itemize}

\begin{table}[t]
\centering
\resizebox{0.8\textwidth}{!}{
\begin{tabular}{||c | c ||} 
 \hline
 Metric & Mathematical Formulation \\ [0.5ex] 
 \hline\hline
 
 Gini Index \cite{ajalloeian2021evaluation} & \Centerstack{ $S(\vec{c}) = 1 - 2\sum_{k=1}^{N} \frac{c_{(k)}}{||\vec{c}||_1} \left(\frac{N - k + \frac{1}{2}}{N} \right) $ \\\\ where $\vec{c} = [c_1  \; c_2 \; c_3]$ \\ and $c_{(1)} \leq c_{(2)} \leq ... \leq c_{(N)}$ is a sorted ordering}\\ 
 \hline
 
 Infidelity \cite{yeh2019fidelity} & \Centerstack{$INFD(\phi, \mathcal{F}, x) = \mathbb{E}_{\mathcal{J} \sim \mu_{\mathcal{J}}} [(\mathcal{J}^T \phi(\mathcal{F}, x) - (\mathcal{F}(x) - \mathcal{F}(x - \mathcal{J})))^2]$}\\ 
 \hline
 
 Sensitivity-n \cite{ancona2017towards} & \Centerstack{ $\sum_{i=1}^{n} \phi_i(\mathcal{F}, x) = \mathcal{F}(x) - \mathcal{F}(x_{[x_S = 0]})$ \\\\ where $x_S = [x_1, ..., x_n] \subseteq x$}\\ 
 \hline
 
\Centerstack{Mean Generalizability \&\\
Relative Consistency \cite{fel2022good}} & \Centerstack{ $MeGe = \left(1 + \frac{1}{|S^=|} \sum_{\delta \in S^=} \delta\right)^{-1}$ \\\\ $ReCo = \max_{\gamma \in S} TPR(\gamma) + TNR(\gamma) - 1 $}\\ 
 \hline
 
 SSR \cite{dabkowski2017real} & \Centerstack{$s(a, p) = \log(\tilde{a}) - \log(p)$ \\ $\tilde{a} = \max(a, 0.05)$ \\\\ where $a$ is the minimum rectangular crop of the salient region}\\ 
 \hline
 
 Robustness-S \cite{hsieh2020evaluations} & \Centerstack{$\epsilon_{x_S}^* = g(\mathcal{F}, x, S) = \{ \min_\delta ||\delta||_p \space s.t. \space \mathcal{F}(x + \delta) \neq y, \delta_{\bar{S}} = 0$ \} \\\\ where $y = \mathcal{F}(x), \bar{S} = U \symbol{92} S$ }\\ 
 \hline
 
 Max Sensitivity \cite{yeh2019fidelity} & \Centerstack{$SENS_{MAX}(\mathcal{I}, \mathcal{F}, x, r) = \max_{||y - x|| \leq r} || \mathcal{I}(\mathcal{F}, y) - \mathcal{I}(\mathcal{F}, x) ||$}\\
 \hline
 
 Relative Stability \cite{agarwal2022rethinking} & \Centerstack{$RIS(x, x', e_x, e_{x'}) = \max_{x'} \frac{||\frac{e_x - e_{x'}}{e_x}||_p}{max(||\frac{x - x'}{x}||_p, \epsilon_{min})}$ \\\\ where $e_x = \phi(\mathcal{F}, x), e_{x'} = \phi(\mathcal{F}, x'), x' \in \mathcal{N}_x$}\\  [1ex]
 \hline
\end{tabular}
}
\caption{Mathematical definitions of different metrics. Section \ref{subsec:evaluate-quality-explanation} provides all the required preliminaries needed to understand the mathematical formulation for a given metric.}
\label{table:metrics-math-definitions}
\vspace{-24pt}
\end{table}

Table \ref{table:metrics-and-properties} provides a summary of the different metrics in terms of the explanation methods evaluated, the datasets on which the evaluations were carried out, and most importantly, the properties that each metric corresponds to. The explanation method in bold, where applicable, is the best-performing method amongst all the evaluated methods on that metric. We can observe the following from Table \ref{table:metrics-and-properties}:

\begin{itemize}[leftmargin=*]
    \item While different methods perform well on different metrics, it can be observed that methods incorporating smoothing or SmoothGrad seem to perform better in general. 
    
    \item We observe that many metrics primarily evaluate fidelity, i.e., how accurately the explanation method reflects the underlying model and its prediction. As mentioned earlier, fidelity is perhaps the most researched metric for explainability. 
    We however notice that there are not many metrics that directly evaluate stability. The metrics that evaluate stability are largely based on measuring the distance between the original explanation maps and the perturbed explanation maps. Evaluating stability by measuring distances between maps is a theme that we shall observe in the rest of this survey, as we look at different distance metrics and their corresponding applications in the context of stability.
\end{itemize}

The latter point is of key interest to us; stability or robustness is a key desired property of a good explanation method. While there are other desired properties (e.g. sparsity is visited in Section \ref{sec:adv-connect}), fidelity and robustness are essential properties. 
Table \ref{table:metrics-and-properties}, however, suggests that the robustness of explanation methods is not yet as well examined as fidelity. A lack of robustness undermines the deployment of an explanation method in a practical setting, as illustrated in Section \ref{sec:intro}. 

Keeping the above points in mind and considering the practical significance of explanation robustness, as well as the potential adverse consequences of a lack thereof, it is thus essential to evaluate where the field of explanation robustness is currently positioned and look at directions it could take moving forward. This domain is currently neither too embryonic in development, like some other properties, nor is it too widely researched and too far developed. Therefore, this serves as an ideal moment to take note of the field as a whole, understand its current status and development, and use that knowledge to provide collated information to increase overall awareness and motivate further research. In the following sections, we shall look at a few generic methods to evaluate robustness. We will then follow it up with a comprehensive discussion on robustness, including different ways to attack and defend explanation methods in DNN models.



\newcolumntype{P}[1]{>{\RaggedRight\arraybackslash}p{#1}}

\begin{table}[t]
\centering
\resizebox{\textwidth}{!}{
\begin{tabular}{||P{3.6cm} | P{6.0cm} | P{2.6cm} | P{2.3cm}||} 
 \hline
 Metric & Explanation Methods & Dataset & Properties \\ 
 \hline\hline
Gini Index \cite{ajalloeian2021evaluation} & Gradient, Smooth Gradient, Uniform Gradient, \textbf{$\beta$-Smoothing} \cite{dombrowski2019geometry} & \textbf{Img:} ImageNet & Sparsity \\ 
 \hline
 
 Infidelity \cite{yeh2019fidelity} & Gradient, SmoothGRAD, Integrated Gradient, \textbf{IG-SG}, Guided BackProp \cite{SpringenbergDBR14}, GBP-SG, SHAP & \textbf{Img:} ImageNet, CIFAR10, MNIST & Fidelity \\ 
 \hline
 
 
 Sensitivity-n \cite{ancona2017towards} & Gradient * Input, Integrated Gradient, $\epsilon$-LRP, DeepLIFT & \textbf{Img:} MNIST, CIFAR10, ImageNet, \textbf{Txt:} IMDB & Completeness, Fidelity\\ 
 \hline
 
 
Mean Generalizability \& Relative Consistency \cite{fel2022good} & Saliency, Gradient * Input, Integrated Gradient, SmoothGrad, GradCAM, GradCAM++, \textbf{RISE} \cite{petsiuk2018rise} & \textbf{Img:} ImageNet & Generalizability \& Consistency \\ 
 \hline
 
 
  SSR \cite{dabkowski2017real} & Max-Bounding Box \cite{dabkowski2017real}, Center Box \cite{dabkowski2017real}, Gradient, Excitation Backprop \cite{zhang2018top}, \textbf{Masking Model} \cite{dabkowski2017real} & \textbf{Img:} ImageNet & Fidelity, Comprehensibility, Efficiency \\ 
 \hline
 
 
 Robustness-S \cite{hsieh2020evaluations} & Gradient, Integrated Gradient, Expected Gradient \cite{erion2019learning}, SHAP, Leave-One-Out \cite{zeiler2014visualizing}, BBMP \cite{fong2017interpretable}, Counterfactual Explanations, \textbf{Greedy-AS} \cite{hsieh2020evaluations} & \textbf{Img:} MNIST, ImageNet, \textbf{Txt:} Yahoo! Answer& Fidelity \\ 
 \hline
 
 
 Max Sensitivity \cite{yeh2019fidelity} & Gradient, SmoothGRAD, ntegrated Gradient, \textbf{IG-SG}, Guided BackProp, GBP-SG, SHAP & \textbf{Img:} ImageNet, CIFAR10, MNIST & Stability \\
 \hline
 
 
 Relative Stability \cite{agarwal2022rethinking} & Gradient, Gradient * Input, Integrated Gradient, \textbf{SmoothGrad}, SHAP, LIME & \textbf{Tab:} Adult, German Credit, COMPAS & Stability \\  [1ex]
 \hline
\end{tabular}
}
\caption{Summary of different quantitative metrics along with explanation methods used for their evaluation, and the datasets on which the evaluations were conducted (\textbf{Img:} image datasets, \textbf{Txt:} text datasets, \textbf{Tab:} tabular datasets). Method in \textbf{bold} is the one that performed the best on that metric. Last column lists the properties that the metric satisfies.}
\label{table:metrics-and-properties}
\vspace{-30pt}
\end{table}
    
    
    

\vspace{-5pt}\section{Evaluating the Robustness of Explanation methods}
\label{sec:evaluate-robustness-explanation}
In Section \ref{subsec:evaluate-quality-explanation}, we observed that stability is generally evaluated by comparing the explanation maps generated for the original input and the perturbed input. In this section, we study the different metrics used for comparing two explanations; a description of these metrics is followed by a brief overview that highlights the practical applications of these metrics in the context of explanation stability. By evaluating the distance between the two explanation maps, we can quantify the robustness of the corresponding explanation method. This section thus serves as a preliminary for the rest of the survey, where we shall look at attributional attacks and attributional robustness, and the role of these distance evaluation metrics, in much greater detail. Some of the metrics that have been proposed are:

\begin{itemize}[leftmargin=*]

    \item \textbf{$L_p$ Distance}: An intuitive and straightforward metric to compare two explanation maps is to compute the normed distance between them. Some of the widely used metrics are median $L_1$ distance, used in \cite{artelt2021evaluating}, and the $L_2$ distance \cite{hubert2021PD, sinha2021NLP, dombrowski2019geometry}. Mean Squared Error (MSE), a metric derived from $L_2$ distance, is also very popular.
    

    \item \textbf{Cosine Distance Metric} \cite{ajalloeian2021evaluation, WangWRMFD20}: Let $z$ and $z'$ be the original and perturbed activation maps respectively. The Cosine Distance (cosd) is defined as $1 - \frac{<z, z'>}{||z||_2||z'||_2}$, where $<z, z'>$ represents the dot product between the two flattened explanation vectors. The metric thus measures the change in direction between the two maps, and a lower cosine distance generally indicates higher similarity.
    

    \item \textbf{SSIM} \cite{arun2020assessing, zhang2022quantifying}: Structural Similarity Index Measure (SSIM) is a metric used to measure the perceptual similarity between two images (the original and perturbed attribution maps in our case). Introduced by \citet{wang2004image}, it is the weighted combination of luminance, contrast, and structure of the two images; a detailed, mathematical definition of the metric can be found in \citet{wang2004image}. The maximum value of SSIM is 1, which corresponds to identical images, whereas a value of 0 indicates no structural similarity.
    
    

    \item \textbf{LPIPS} \cite{ajalloeian2021evaluation, zhang2018unreasonable}: Learned Perceptual Image Patch Similarity (LPIPS) is used to, as the name suggests, measure the perceptual distance between two images (the original and perturbed attribution maps in our case). In particular, it captures the distance between the internal activations of the respective images on a pre-defined network; a lower LPIPS value indicates higher similarity between the images. A more detailed description of the metric can be found in \citet{zhang2018unreasonable}.
    

    \item \textbf{Spearman's Rank-Order Correlation} \cite{fel2022good, WangWRMFD20}: Spearman's Rank-Order Correlation Coefficient (Spearman's $\rho$) is a non-parametric metric used to measure the strength of the relationship between two ranked variables. For explanation methods, it is used to compare two explanation maps (generally between the original and perturbed explanation maps). For example, given a sample size $n$, the $n$ raw scores $X_i, Y_i$ are converted to rank vectors $R(X_i), R(Y_i)$ and Spearman's Correlation Coefficient $r_s$ is computed as the Pearson Correlation Coefficient between the rank variables i.e., $r_s = \frac{cov(R(X), R(Y))}{\sigma_{R(X)}\sigma_{R(Y)}}$. Other similar metrics included are Pearson's Correlation Coefficient (PCC), which is a measure of the linear correlation between two vectors, Kendall's Rank-Order Correlation (Kendall's $\tau$), which is also used to measure the rank-order correlation of two explanation maps, and top-$k$ intersection \cite{WangWRMFD20}, which computes the intersection between the $k$ features with highest attributions in the two given maps. Since features with higher rank in the attribution map are generally regarded as more important and contribute the most to the model prediction, these are good metrics to compare the similarity of two explanation maps.   
    

    

    

    \item \textbf{Prediction-Saliency Correlation (PSC)} \cite{zhang2022quantifying}: \citet{zhang2022quantifying} proposed the Prediction-Saliency Correlation (PSC) metric to capture the correlation between the changes in model predictions and the changes in the corresponding saliency maps. Quantitatively, each image $x_{i}, i \in \{1, 2, ..., N\}$ in a test set has an associated prediction $p_{i}$ and saliency map $m_{i}$. Each image is also perturbed to generate a perturbed image $x_{i}^{'}$ with prediction $p_{i}^{'}$ and saliency map $m_{i}^{'}$. $d_{i}^{p}$ is defined as Jensen-Shannon (JS) divergence between $p_{i}$ and $p_{i}^{'}$, and $d_{i}^{m}$ is the JS divergence between $m_{i}$ and $m_{i}^{'}$. Finally, PSC is defined as the Pearson correlation coefficient between the sets $D_{p} = \{d_{i}^{p}\}_{i=1}^N$ and $D_{m} = \{d_{i}^{m}\}_{i=1}^{N}$, which thus evaluates the correlation between the prediction changes and saliency map changes.
    

    \item \textbf{Top$_{J}$ Similarity} \cite{messalas2019model}: Surrogate models are models constructed to be interpretable and are designed to emulate the behavior of a black-box model, to provide explanations for the black-box model's decisions. \citet{messalas2019model} introduced a metric to compare the explanations of the original model and the surrogate model based on their SHAP values. In particular, let the top $j$ features of the models (ordered by the absolute SHAP values \cite{LundbergL17} be placed in ORIG$_j$ and SUR$_j$, and let $common$ be an array such that $common_i = $ ORIG$_j(i) \cap $ SUR$_j(i), \forall i \in N$, where $N$ is the number of instances. Then, Top$_{j}$ Similarity $ = \frac{avg(common)}{j}$. One can see that this can be generalized for comparing any two specific explanation maps generated by the same method.

\end{itemize}

These metrics provide different ways to compare two different explanations generated by the same method. Armed with this knowledge, a natural follow-up would be to ask about the practical applications of these metrics in the context of explanation methods and stability. A brief overview can be provided as follows:

\begin{itemize}[leftmargin=*]
    \item The \textit{$L_2$ norm} is a metric that is both generic as well as highly intuitive, and is widely used to evaluate the distance between two explanations.
    
    \item For evaluating attributional attacks, \textit{Spearman's rank order correlation} is a popular metric. This metric is used frequently for targeted and COM-shift based attacks (refer to Section \ref{sec:attack-explain} for further details). \textit{top-$k$ intersection} is another popular metric that is used in the eponymous Top-$k$ fooling (refer to Section \ref{sec:attack-explain}).
    
    \item For attributional robustness, i.e., methods for improving the robustness of explanation methods, \textit{Kendall's $\tau$} and \textit{top-$k$ Intersection} are the most frequently used metrics.
    
    \item The popularity of \textit{Spearman's $\rho$}, \textit{Kendall's $\tau$} and \textit{top-$k$ Intersection} can be attributed to the fact that they concern themselves with the correlation between the feature rankings, and not the actual attribution values themselves. In other words, these metrics compare the set of most relevant features from either map and evaluate the intersection or correlation between these sets.
    
    \item Some other metrics used are \textit{ROAR}, \textit{Energy Ratio}, and the \textit{Pixel-wise Sum of Squared Distances}, but these metrics have not found wide usage.
    
\end{itemize}

In the following sections, we shall see how these metrics are practically applied; in particular, we shall see their utility in evaluating attributional attacks and attributional robustness.


\vspace{-5pt}\section{Attacking Explainability Methods} \label{sec:attack-explain}

Post-hoc explainability methods play an essential role in comprehending the decisions made by a neural network model. The term ``attack" has been loosely used in multiple research papers, and it is necessary to build a concrete definition to understand the sections as we advance. An attack in this context means any manipulation which makes the model deviate from its primary objective and produces undesirable changes in the output. This section aims to categorize how manipulations can be applied to a classification system and how these changes are quantified.

Let $x \in \mathcal{R}^d$ be the input data and $y \in Y = \{1,...,k\}$ be the labels assigned to the classes of data. Let $\mathcal{D}$ be the data distribution from which the sample $(x,y) \in \mathcal{D}$ is drawn and let the classifier be $\mathcal{F} : x \rightarrow y$. An \textit{Adversarial attack} is defined as a perturbed input $x'$ corresponding to input x such that $\mathcal{F}(x) != \mathcal{F}(x')$. Let $\mathcal{I}(x)$ be the attribution map obtained for input $x$ using a method $\mathcal{I}$ \cite{SimonyanVZ13,erhan2009visualizing,SelvarajuCDVPB17,SundararajanTY17,SmilkovTKVW17,ShrikumarGK17,DattaSZ16,StrumbeljK14,SpringenbergDBR14,PengM21a}. Then an \textit{Attributional attack} is defined as perturbing the input $x$ to obtain $x''$ such that $\mathcal{F}(x) = \mathcal{F}(x'')$ but $\mathcal{C} \{\mathcal{I}(x) != \mathcal{I}(x'')\}$, where $\mathcal{C}$ is the similarity metric like top-$k$ intersection, Spearman's ($\rho$) or Kendall's ($\tau$) rank correlation metrics. Similarly, a \textit{Trust attack} is given by $x'''$ obtained from x such that $\mathcal{F}(x) != \mathcal{F}(x''')$ but $\mathcal{C}\{\mathcal{I}(x) = \mathcal{I}(x''')\}$.

\begin{wrapfigure}[13]{r}{0.66\textwidth}
\vspace{-12pt}
\centering
\begin{subfigure}{.21\textwidth}
  \includegraphics[width=1.0\textwidth, height=0.18\textheight]{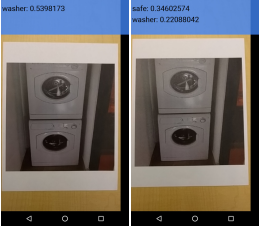}  
  \subcaption{Adversarial Attack}
  \label{attack-types-adv}
\end{subfigure}
\begin{subfigure}{.22\textwidth}
  \includegraphics[width=1.0\textwidth, height=0.185\textheight]{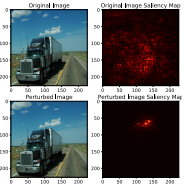}  
  \subcaption{Attributional Attack}
  \label{attack-types-attr}
\end{subfigure}
\begin{subfigure}{.22\textwidth}  
  \includegraphics[width=1.0\textwidth, height=0.18\textheight]{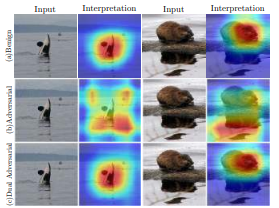}  
  \subcaption{Trust Attack}
  \label{attack-types-trust}
\end{subfigure}
\vspace{-12pt}
\caption{(a) \cite{AkhtarM18} (left) clean image (right) adversarially perturbed image (b)\cite{ghorbani2017fragile} (top) clean image (bottom) attributionally perturbed image (c)\cite{zhang2018fire} Explanation maps are same yet different predictions.}
\label{attack-types}
\vspace{-14pt}
\end{wrapfigure}

 Thus, there are three kinds of attack scenarios possible (for example, by adding a visually imperceptible perturbation to an image): 
\begin{enumerate}[leftmargin=*]
    \item Adversarial attack (Figure \ref{attack-types-adv}) leads to different predictions and explanations. 
    \item Attributional attack (Figure \ref{attack-types-attr}) maintains the same predictions but leads to different explanations. 
    \item Trust attack (Figure \ref{attack-types-trust}) leads to different predictions but similar explanations. 
\end{enumerate} 
Figure \ref{attack-types} gives an example for each type of attack. Ideally, robustness methods (discussed in the next section) should focus on achieving the fourth combination, viz. obtaining exact predictions and similar explanations under perturbed inputs to the model. All three attacks pose critical security concerns to DNN models, yet only adversarial attacks are a well-researched topic. We hence focus on attributional attacks in this work and touch upon a few trust attacks as we advance. The decisions made by a model to arrive at a prediction could be wrong despite a correct prediction, and attributional attacks show this; hence it is imperative to understand them. The primary attacks in this context are focused on: 
\begin{inparaenum} \item Image\cite{ghorbani2017fragile} - pixels of an image are perturbed at specific locations; \item Text/NLP (Natural Language Processing)\cite{slack2019posthoc,ivankay2022text} - words are replaced by their synonyms in a sentence; and \item Tabular data (ML models) \cite{hubert2021PD} - dataset is poisoned to conceal the suspected behavior (bias) \end{inparaenum}.

\begin{wrapfigure}[7]{r}{0.4\textwidth}
    \vspace{-16pt}
    \centering
    \includegraphics[width=0.35\textwidth]{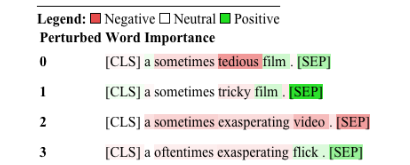}
    \caption{Example of attributional attack on text data (reproduced from \citet{ivankay2020far})}
    \label{fig:text-attack}
\end{wrapfigure}

\vspace{-6pt}\subsection{Common Techniques and Metrics}
\label{subsec:common-tech}
Deep neural network models are currently used as black boxes due to their high accuracy for various tasks. Recently, efforts have been made to understand the decisions taken by these models. There have been efforts to explain their predictions and an equal amount of effort to show how inherently ``fragile" they are, along with defenses to fix it. After \citet{ghorbani2017fragile} showed that current gradient methods are vulnerable to an attack, others researched for approaches to attack other data types like text/NLP \cite{slack2019posthoc,ivankay2022text} (Figure \ref{fig:text-attack}) and tabular data \cite{hubert2021PD}.
Below we list a few standard techniques considered while attacking models, which have successfully proved the vulnerability\cite{ghorbani2017fragile} of current explanation methods.

\subsubsection{Top-$k$ fooling}
Top-$k$ represents the ranked set of pixels of high importance in an explanation map. This attack aims to reduce the scores in explanations corresponding to those pixels that initially had the highest $k$ values. This technique applies to all three data spaces: scores representing pixel values in explanation maps for images, weightage of words for text/NLP, and feature scores for tabular data. The size of the intersection of the topmost $k$ essential features before and after perturbation can quantitatively measure the technique's effect. Top-$k$ is one of the most frequently used attack techniques \cite{ghorbani2017fragile,Chen0RLJ19,SarkarSGB21}.

\subsubsection{COM shift}
The Center Of Mass (COM) stands for the center of feature importance mass. In a COM shift attack, there is an effort to deviate the explanations as much as possible from the original center of mass. It is mainly applicable for attacking image data. A specific example of a normalized feature importance function $\mathcal{I}(x)$\cite{ghorbani2017fragile} is an explanation for an image $x$ of size $\mathcal{W} \times \mathcal{H}$, the center of mass is defined as $\mathcal{C}(x) = \Sigma_{i\in\{1,...,\mathcal{W}\}} \Sigma_{j\in\{1,...,\mathcal{H}\}} \mathcal{I}(x)_{i,j}[i,j]^\mathcal{T}$. \citet{ghorbani2017fragile} use the loss function $||\mathcal{C}(x) - \mathcal{C}(x_{t})||_{2}$ to construct a COM shift based attack on the explanation. The effectiveness of the attack can be measured through other metrics like $L_1$ norm\cite{heo2019advmodel} as well.

\begin{wrapfigure}{r}{0.35\textwidth}
    \vspace{-16pt}
    \centering
    \includegraphics[width=0.3\textwidth]{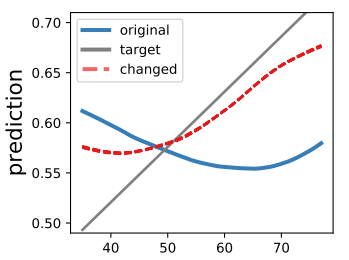}
    \caption{\citet{hubert2021PD} construct a targeted Attack on tabular data by data poisoning (equivalent to input perturbation for images) which is captured by the deviation (red line) from the  original dependency graph (blue line) can be deviated towards a target graph.}
    \label{fig:targeted-attack}
    \vspace{-19pt}
\end{wrapfigure}

\subsubsection{Targeted attack}
Targeted attack refers to intentionally making the explanation methods generate false explanations which resemble a specific attribution map corresponding to an incorrect label. Unlike previous attacks, we have a target map which is used for evaluation. One approach \cite{ghorbani2017fragile} aims to generate false explanations by projecting unimportant input parts to be crucial in obtaining predictions. This technique is mainly used to attack images, where unimportant parts represent objects in images and a set of features in tabular data. Figure \ref{fig:targeted-attack} gives an example of how the original dependency graph (for tabular data) could deviate towards a target graph via data poisoning\cite{hubert2021PD}. The effectiveness is mainly measured through the $L_{2}$ norm. But other metrics like PCC (Pearson Correlation Coefficient), SSIM, Spearman's rank order correlation, etc., are used for image data.

\vspace{-6pt}\subsection{Attributional Attack Methods}
In this section, we broadly categorize existing attributional attack methods based on what information is accessible to an attacker. Section \ref{subsubsec:model-specific} focuses on attacks being constructed specifically to the model, while Section \ref{subsubsec:model-agnostic} focuses on attacks built without access to the inner workings of the model. This categorization would give us insights into the intuition behind formulating the attacks.

\subsubsection{Model-Specific Attacks}
\label{subsubsec:model-specific}
Model-specific approaches are based on the specific structures of the deep learning model, which are exposed or made available to an attacker. IG, Simple Gradients, Attention Maps, Saliency Maps, etc., fall under model-specific explainability methods. The algorithmic approach used to construct attacks on such methods is to iteratively craft a perturbed input that modifies the intermediate layers of the neural network, thereby affecting the explanations but maintaining the logits of the final layer to ensure prediction stability.

\noindent \paragraph{Image Data}
\citet{ghorbani2017fragile} proposed the first attributional attack which satisfies the definition with example in Figure \ref{attack-types-attr}. The attack includes a set of techniques called Iterative Feature Importance Attacks(IFIA), which add a small imperceptible noise within the $\epsilon$-budget that does not change the prediction of the network but changes the explanation maps significantly under metrics like top-$k$ intersection and Spearman's rank correlation. They found an intriguing observation that the current explanation methods are so fragile that explanations are susceptible to even a natural baseline using random sign perturbation.

While the IFIA method by \citet{ghorbani2017fragile} is largely an untargeted attack. \citet{dombrowski2019geometry} propose a targeted attack. The attack shows an example where an input dog image can be perturbed to obtain explanations close to a cat. Specifically, given a $x_{adv}$ as the adversarial input, $\mathcal{I}(x)$ as the explanation map, $\mathcal{I}^{t}$ as the target explanation map, $\mathcal{F}(.)$ as the output of the network, the method optimizes the loss $\mathcal{L} = || \mathcal{I}(x_{adv}) - \mathcal{I}^{t}||^{2} + \gamma ||\mathcal{F}(x_{adv}) - \mathcal{F}(x)||^{2}$.

\noindent \paragraph{Text Data}
\citet{ivankay2022text} proposed a similar attack in NLP domain. Initially, importance ranking $\mathfrak{I}$ for each token $i$ in the input sample $s$ is calculated. Specifically, for a text classifier $\mathcal{F}$, labels $l$ and distance metric $d$ with $s_{w_{i}}\rightarrow0$ denoting the $i$-th word being masked to the zero embedding token, the importance ranking is defined as $\mathfrak{I}_{w_{i}} = d[A(s_{w_i \rightarrow 0}, \mathcal{F}, l), A(s, \mathcal{F}, l)]$. For each word $w_{i}$ in the sample, a set of substitution candidates is extracted (using the counter-fitted GloVe synonym embeddings). The method generates the perturbed sample by replacing each word with its synonym, which maximizes the distance between the attributions. Unlike images, the text perturbations can be visually perceived(Figure \ref{fig:text-attack}), which is why the method makes sure the semantic similarity is also maintained by limiting the word replacements and applying Part-of-Speech (POS) filters.

\subsubsection{Model-Agnostic Attacks}
\label{subsubsec:model-agnostic}
Model-agnostic approaches do not consider the model's structure and can be applied to any deep learning(DL) model. LIME, SHAP etc., fall into this category. The standard baseline for attacks on such methods is to intuitively create an adversarial model that would fool the explanation methods. A significant difference between these attacks and the model-specific attacks is that the input is perturbed to affect the explanations in model-specific attacks. In contrast to model-specific attacks, the input is perturbed to understand how explanations are calculated in model-agnostic attacks. This information is later used to build an adversarial classifier that fools the explanation methods in model-agnostic attacks.

\noindent \paragraph{Image Data}
\citet{slack2019posthoc} proposed an attack on post-hoc explanation techniques like LIME and SHAP. The core logic of LIME and SHAP is to explain individual predictions of a given black box model by constructing local interpretable approximations (e.g., linear models). Each such local approximation is designed to capture the behavior of the black box within the neighborhood of a given data point. These neighborhoods constitute synthetic data points generated by perturbing features of individual instances in the input data. However, samples generated using such perturbations could potentially be off-manifold or out-of-distribution (OOD). The attack is intuitively constructed on this information.

An OOD classifier is trained on the mixed dataset (original samples are set with the label False while perturbed samples are set with the label True). The adversarial model is constructed in such a way that it behaves in a biased manner (making predictions based on sensitive attributes like race, gender, etc.) on the original samples while behaving unbiasedly on the OOD samples. This ensures that LIME/SHAP are fooled when they try to understand the black box model with perturbed samples. The bottleneck for these attacks lies in how accurately the OOD classifier can identify a perturbed sample.

\noindent \paragraph{Tabular Data}
\citet{hubert2021PD} proposed an attack on partial dependence (PD) explanation method via data poisoning. At its core, PD presents the expected value of the model's predictions as a function of a selected variable. Specifically, given a model $\mathcal{F}$ and a variable $c$ in a random vector $X$ and $X_{-c}$ representing random vector X, where $c$-th variable is replaced by value $z$, $\mathcal{P}\mathcal{D}_{c}(\mathcal{X},z) := E_{\mathcal{X}_{-c}}[f(\mathcal{X}^{c|=z})]$. The attack uses a genetic algorithm that focuses on iteratively perturbing the values of the individuals in the dataset after each iteration through crossover, mutation, evaluation, and selection techniques. Crossover swaps columns between samples to produce new ones, mutation adds gaussian noise to the samples, evaluation calculates the predefined loss function, which is the $L_{2}$-norm between current PD and target PD, and selection reduces the number of samples in the total population using rank selection. None of these techniques use the information on the inner working of models in line with the definition of model-agnostic attacks.

Other important details related to the attack spaces, performance metrics, and techniques used within these various attack methods have been summarized in the table \ref{table:attr-attacks}. It can be observed that there are certain combinations of explanation methods and attack techniques (\ref{subsec:common-tech}) which have not been tried out yet.

\newcolumntype{P}[1]{>{\RaggedRight\arraybackslash}p{#1}}

\begin{table}[ht]
\centering
\resizebox{\textwidth}{!}{
\begin{tabular}{||P{3cm} | P{2.3cm} | P{5cm} | P{4.5cm} | P{3.4cm}||} 
 \hline
 Reference & Technique Used & Evaluation Metric & Explanation Methods & Dataset \\ [0.25ex] 
 \hline\hline
 
\citet{ghorbani2017fragile} & COM, Top-$k$, Target & Spearman's $\rho$, Top-$k$ intersection & Simple Gradient, DeepLIFT, IG & ImageNet, CIFAR-10  \\ 
 \hline
 
\citet{slack2019posthoc} &  Top-$k$ & Top-$k$ intersection & LIME, SHAP & COMPAS by ProPublica, Communities and Crime, German Credit \\
 \hline
 
 \citet{hubert2021PD} & Target & $L_{2}$-Norm/MSE & Partial Dependence & Friedman's work, UCI Heart dataset  \\ 
 \hline
 
 \citet{sinha2021NLP} & Top-$k$ & Spearman's $\rho$, Top-$k$ intersection, $L_{2}$-Norm/MSE  & IG, LIME  & AG News, IMDB, SST-2  \\ 
 \hline
 
 \citet{ivankay2022text} & Top-$k$ & Spearman's $\rho$, Top-$k$ intersection, PCC & Saliency Maps, IG, Attention mechanisms & AG News, MR reviews, IMDB, Fake News, Yelp \\ 
 \hline
 
 \citet{dombrowski2019geometry} & Target & SSIM, $L_{2}$-Norm/MSE, PCC & Gradient, GradientxInput, IG, Guided BackProp, LRP, Pattern Attribution  & ImageNet, CIFAR-10  \\ 
 \hline
 
 \citet{heo2019advmodel} & COM, Top-$k$, Target & Spearman's $\rho$, Top-$k$ intersection &  LRP, Grad-CAM, SimpleGrad & ImageNet \\ 
 \hline
\end{tabular}
}
\caption{Summary of different attributional attacks along with the explanation methods used for their evaluation, and the datasets on which the attacks were performed.}
\label{table:attr-attacks}
\vspace{-6mm}
\end{table}

\vspace{-6pt}\subsection{Other Types of Attacks}
In the ensuing section, we present attacks from a few other dimensions that could not be captured in earlier sections, given the definition of an attack we considered. Strictly speaking, these attacks can't be termed attributional attacks, yet the techniques are worth knowing to formulate innovative attacks in the future.\\

\subsubsection{Model-Specific attack on Model-Agnostic Explanation Methods} 

\noindent \paragraph{Text Data}
\citet{sinha2021NLP} proposed an attack on model-agnostic methods using an innovative approach of employing traditional techniques in model-specific attacks. The idea is similar to \citet{ivankay2022text}, where the ranking of words is initially calculated through the leave-one-out approach. Loss functions (Location of Mass (LOM) and $L_{2}$-norm) are used as metrics to measure the closeness of explanations. In decreasing order of importance, each word is perturbed after each iteration with its synonym (out of $k$ closest embeddings) that yields the highest loss.
This attack focuses on IG (model-specific explanation) and LIME (model-agnostic explanation) methods. The attack treats the output of LIME as an explanation in itself and tries to shift it as far away from the original explanation as possible through input perturbations. Despite the attack being focused in a model-specific fashion, it also yields promising results on model-agnostic methods.
This raises an important question we could try out the reverse fashion, i.e., employing model-agnostic techniques for model-specific methods. 

\vspace{6pt}
\subsubsection{Attack on Feature space}
\vspace{-3pt}
\noindent \paragraph{Image Data}
\citet{heo2019advmodel} proposed an attack that fools a neural network by adversarial model manipulation without hurting the accuracy of original models (model fine-tuning step). It is similar to the techniques used for attacking model-agnostic explanation methods where we ensure the model behaves in such a way that the explanation method is fooled.
The attack achieves this by incorporating an additional loss function (specific to the technique used) in the penalty term of the objective function for fine-tuning. Consider a top-$k$ fooling attack given a dataset $\mathcal{D}_{fool}$, a neural network $w$, a heatmap generated by an explanation method $\mathcal{I}$ for $w$ and class c denoted by $h_{c}^{\mathcal{I}} = \mathcal{I}(x,c;w)$ and $\mathcal{P}\textsubscript{i,k}(w_{0})$ be the set of pixels that had the top-$k$ highest heatmap values for the original model $w_{0}$, then for the $i$-th data point, the additional term in loss function is $\mathcal{L}_{\mathcal{F}}^{\mathcal{I}}(\mathcal{D}_{fool};w,w_{0}) = \frac{1}{n} \sum_{i=1}^{n}\sum_{j\in\mathcal{P}_{i,k}(w_{0})}|h_{y_{i},j}^{\mathcal{I}}(w)|$.
Just as we have accuracy for the predictions, this method also defines a quantitative metric called Fooling Success Rate (FSR)\cite{heo2019advmodel} for measuring how well the explanation method has been fooled. A critical analysis of this method gives us insights that it is essentially perturbing the space of model parameters instead of the input space. There could be other spaces, like model architecture, which could be perturbed to generate other innovative attacks.

\subsubsection{Attack on Trust}
Given two almost identical explanations, if their predictions or functionality of models are different, then it is said to be an attack on trust. This branch of attacks goes a step forward to attributional attacks and exploits the basic cues a user takes in understanding the explanations. It is important to note that an attributionally robust model might not be trust-wise robust and vice-versa.

\noindent \paragraph{Tabular Data}
\citet{lakkaraju2020fool} proposed an attack that manipulates the user trust in black box models. The method provides a theoretical framework for understanding and generating misleading explanations and carries out a user study with domain experts to demonstrate how these explanations could mislead the users. User trust, being an abstract term, must be quantified. Precisely, the feature space can be decomposed into $X = X_{D} \times X_{A} \times X_{P}$, where $X_{D}$ corresponds to the desired features D that the user expects to be included, $X_{A}$ corresponds to the ambivalent features A for which the user is indifferent about whether they are included, and $X_{P}$ corresponds to the prohibited features P that user expects to be omitted.

An accepted explanation is one where desired features are included, and the prohibited features are omitted. The same applies to an accepted black box model as well. The method constructs attack using the Model Understanding through Subspace Explanations (MUSE) framework where its objective is to produce potential misleading explanations for an unacceptable black box model, i.e., the explanations contain desired features and ensure prohibited features are not included while the black box model uses the prohibited features to make predictions. This ensures that the users (domain experts) are fooled by accepting the black box model based on misleading explanations, which could cause a potential threat in real-life situations.

\noindent \paragraph{Image Data}
\citet{zhang2018fire} proposed a similar attack on image data. This method aims to show that the existing interpretable deep learning systems (IDLSes) are highly vulnerable to adversarial manipulations. It presents a new class of attacks ADV that ensure the predictions of deep neural networks (DNNs) are different while the explanations are as close as possible. Specifically, given a DNN $\mathcal{F}$, its coupled interpreter $\mathcal{I}$, adversarial input $x^{*}$ generated by modifying a benign input $x^{o}$, ADV ensures the following, \begin{inparaenum} \item $x_{*}$ is misclassified by $\mathcal{F}$ to a target class $c_{t}$, $\mathcal{F}(x_{*}) = c_{t}$ \item $x_{*}$ triggers $\mathcal{I}$ to generate a target attribution map $m_{t}$, $\mathcal{I}(x_{*},\mathcal{F}) = m_{t}$ \item The difference between $x_{*}$ and $x_{o}$, $\delta(x_{*},x_{o})$ is imperceptible \end{inparaenum} 

The above trust attacks give a new dimension to attacks on explanations, as the traditional attributional attack approach manipulates explanations to ensure that the predictions remain the same. This new approach helps us raise concerns about the validity of explanations and how well they are in line with the predictions made. The new dimension covered in these trust attacks is to look at the inverse of the approach of the attributional attack, which should also be tested before the deployment of models. The dual testing would bring out potential biases in the model because it is impossible for a model to be trained in both ways (similar explanations leading to similar predictions and similar predictions leading to similar explanations).

Finally, we understand that existing models are fragile to simple and intuitive attributional attacks. Every attack can be viewed as a path taken in a decision tree, with decisions being various techniques (COM shift, top-$k$), multiple spaces (input, feature), different explanation methods (IG, LIME), and input types (Image, NLP). There are yet many paths that have not been discovered and experimented with. These attacks raise questions over a model's social acceptability; hence, a model must explain itself and be robust in the face of such attacks. In the following section, we shall learn about ways in which we can go about achieving this robustness.

\vspace{-5pt}\section{Ensuring Robustness of Explainability Methods} \label{sec:ensure-robust}

Section \ref{sec:attack-explain} defined multiple notions of an attack for a classifier $\mathcal{F}$. This section focuses on understanding the shortcomings of deep learning networks, which make them vulnerable to attacks, and methods of defense against these attacks. Section \ref{subsec:attack-intuition} focuses on the reasoning behind the success of an attribution attack as, given in Figure \ref{attack-types-attr}. Section \ref{subsubsec:math-approach} discusses the mathematical properties capturing robustness. Sections \ref{subsubsec:common-tech} and \ref{subsubsec:saliency-maps} discuss various types of defenses built toward these attacks, where Section \ref{subsubsec:common-tech} discusses various regularization techniques based on new loss functions and augmentations, and  Section \ref{subsubsec:saliency-maps} focuses on how tweaking the attribution map generation process can provide better tolerance to attributions attacks.

\vspace{-6pt}\subsection{Why Attributional Attacks Work?}\label{subsec:attack-intuition}

One prominent reason attribution attacks prevail is because of the inherent non-linear property of neural network models by which the gradient can change rapidly within small distances of the input space \cite{ghorbani2017fragile,dombrowski2019geometry,WangWRMFD20}. Most explanation methods rely on gradients; hence this is an important issue. This intuition was used to explain the vulnerability to \citet{ghorbani2017fragile} attack (which is one of the benchmarks for testing attributional robustness), showing that the decision boundary of deep neural networks is complex; hence, a slight perturbation in the input would take the model into a different loss contour. 

\begin{wrapfigure}{R}{0.66\textwidth}
\vspace{-10pt}
\centering
\begin{subfigure}{.65\textwidth}
  \includegraphics[width=.49\textwidth, height=0.5\textwidth]{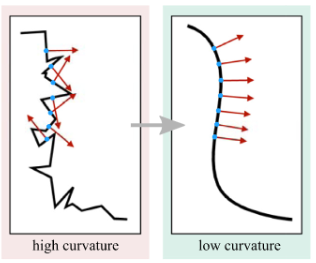}  
  \includegraphics[width=0.49\textwidth, height=0.5\textwidth]{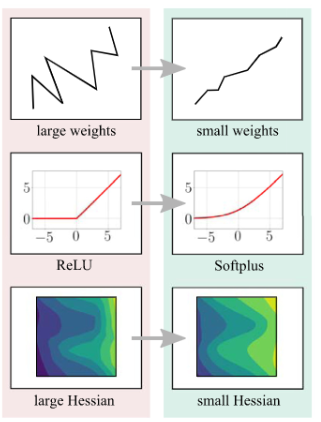}  
\end{subfigure}
\newline
\begin{subfigure}{.65\textwidth}  
  \includegraphics[width=1.0\textwidth]{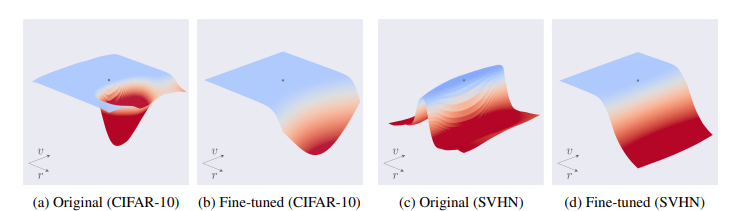}  
\end{subfigure}
\vspace{-3pt}
\caption{(top) \protect \citet{dombrowski2019geometry} provide the intuition for their approach by observing the normals at the decision  boundary. (top left) The gradient (red arrows) changes drastically when moving along a line with high curvature but changes gradually when the curvature is low. (top right) They also propose techniques like weight decay that flattens the angles between piece-wise linear functions, softplus smooths out the kinks of the ReLU function and hessian minimization reduces the curvature locally at the data point. (bottom) \protect \citet{Moosavi-Dezfooli19} shows the benefits of adversarial training by illustrating the negative loss function along normal and random directions r and v. They observe that the original network (bottom: a,c) has large curvature in those directions. Adversarial training results in a lower curvature of the loss (bottom: b, d). The original sample is illustrated with a blue dot along with a light blue surface (classification region of the sample), and a red region (adversarial region), respectively.}
\vspace{-15pt}
\label{smooth-bound}
\end{wrapfigure}

Another perspective to the understanding is by \citet{dombrowski2019geometry,Moosavi-Dezfooli16}, who attribute the vulnerability to the non-smoothness of contemporary neural network models. In Figure \ref{smooth-bound} (top left), we observe the gradient (red arrows) changes drastically while moving along a line with high curvature but more gradually when the curvature is low. 
This vulnerability of explanation methods is related to the large curvature of the output manifold of the neural network. When we perturb the input by retaining the exact prediction, it means that we are moving on a hypersurface S with a constant network output. This attack focuses on explanation methods based on gradients in image space which depend on the class with the highest final score. It can be concluded that the gradient for these methods would be of co-dimension one; hence, its direction would be normal to the hypersurface. After thorough reasoning, we can state that since slight perturbation leads to considerable change in explanations (indirectly the gradients), the curvature of S is large. Similar studies are well-known and used to study the impact of adversarial training on the loss of neural networks, as seen in Figure \ref{smooth-bound}(bottom).


While the previous approaches relied on curvature analysis to explain the vulnerability of networks to attributional attacks, \citet{EtmannLMS19} use the alignment or misalignment between the image and its saliency map to quantify robustness. They point out that robustness induces a larger distance between the data point and the decision boundary, increasing with distance to its closest decision boundary, thereby affecting the corresponding alignment. We observe this approach as simple to indicate a network's vulnerability to attributional attacks and show when such attacks could work.

We discussed why an attack works above, and in the next section, we discuss various types of defenses. 

\vspace{9pt}
\subsection{Principled Approaches}
In this section, we primarily focus on fundamental principles that help exploit the underlying mathematical principles which help obtain the defenses discussed in subsequent sections.
\label{subsubsec:math-approach}

\subsubsection{Certified Attributional Robustness} 
 
Attributional attacks are judged by the change they can make in the original attribution map by perturbing the input within a certain perturbation budget. While finding defenses against such attacks, it is imperative to have ones that provide some guaranteed behavior. Bounds on mathematical properties are one method to give a lower/upper limit on such behavior, which help in knowing how much change we can expect from a system under attack. Deriving bounds and limits for attributional robustness can be done by various methods as below.

\indent
\citet{dombrowski2019geometry} have shown how geometry can explain the manipulations of explanations and derived an upper bound on the maximal change possible for a gradient map. They propose that the Softplus function can replace the ReLU function to reduce the non-smoothness of the network to achieve a stable gradient saliency map. Similarly, \citet{HuaiLMYZ22} also viewed ensuring robustness from the angle of reducing the change caused to explanations because of an attack. They aimed to minimize an upper bound on the worst-case loss caused by any adversarial attack. Thus, they certified robustness by adding a bounding-based regularization term to its loss function, which minimizes the upper bound on the maximum difference between any two explanations for inputs perturbed within a norm ball.
\citet{SinghKMSBK20} took a different approach to the bounding methodology where they aimed to minimize the upper bound for the spatial correlation function between the input image and the explanation map. They show attributional vulnerability, which is the maximum possible change in the gradient-based feature importance score is upper bounded by the maximum of the distance between $\mathcal{I}^y (x + \delta)$ and $x + \delta$ for $\delta$ in the neighborhood of $x$. It is derived that $\left.\max _{\delta \in B_\epsilon} \| \mathcal{I}^y(x+\delta)-\mathcal{I}^y(x)\right)\left\|\leq 2 \max _{\delta \in B_\epsilon}\right\| \mathcal{I}^y(x+\delta)-(x+\delta)\|+\| \epsilon \|$. Where ${\mathcal{I}^y}$ is the gradient-based feature importance score. Further, triplet loss formulation minimizes the distance between two quantities.

\subsubsection{Lipschitz Continuity} \label{subsubsec:liptschitz-conti}

Lipschitz continuity of a function limits the speed at which the function changes. Concerning our use case, this property ensures stability by measuring changes in the outputs relative to the changes in the inputs. By definition, Lipschitz continuity is a global property, as it captures the deviations throughout the input space. But for robust interpretability, we are only concerned with how much the outputs change with small perturbations in the inputs, also called neighboring inputs. Thus, \citet{alvarez2018robustness} propose to rely on the point-wise, neighborhood-based local Lipschitz continuity. Local stability is again defined in either the continuous or weaker empirical notions for discrete finite-sample neighborhoods. The robustness of a saliency map for an input depends strongly on the local geometry, so increasing the robustness translates into smoothening this geometry, naturally captured by the local-Lipschitz property. \citet{WangWRMFD20} effectively use this property to characterise robustness to attacks by \citet{ghorbani2017fragile}. They highlight the usefulness of this property and its intuitive application in training in post-hoc techniques. \citet{WangWRMFD20} formally define local Lipschitz continuity and global Lipschitz continuity and use it to define attributional robustness as follows:

\noindent \textit{\underline{Definitions:} Lipschitz Continuity :} A general function $h : \mathbb{R}^{d_{1}} \rightarrow \mathbb{R}^{d_{2}}$ $(\lambda, \delta_{p})$-locally robust (\cite{WangWRMFD20}[Definition 5]) if $\forall x' \in B(x, \delta_{p}), ||h(x) - h(x')||_{p} \leq \mathcal{L}||x - x'||_{p}$ . Similarly, h is L-globally Lipschitz continuous  if $\forall x' \in \mathbb{R}^{d_{1}}, ||h(x) - h(x')||_{p} \leq L||x - x'||_{p}$. \textit{Attributional Robustness with respect to Lipschitz Continuity:} An attribution method $\mathcal{I}(x)$ is $(\lambda, \delta_{p})$-locally robust (\cite{WangWRMFD20}[Definition 6]) if $\mathcal{I}(x)$  is $(\lambda, \delta_{p})$-locally Lipschitz continuous, and $\lambda$-globally robust if $\mathcal{I}(x)$ is $\lambda$-globally Lipschitz continuous.

\citet{WangWRMFD20} connect robustness with conditioning the Lipschitz continuity property, which bounds the models' gradients to obtain the smoothness of the model's decision surface. They exploit this connection to construct a regularizer to enforce the property. They discuss a stochastic smoothing of the local geometry called Uniform Gradient, which is defined as $g(\mathbf{x})=\mathbb{E}_p [\nabla_{\mathbf{x}} \mathcal{F}(\mathbf{z})]$ where $p(\mathbf{z})=U(\mathbf{x}, r)$ is a uniform distribution.
\citet{AgarwalJAU0L21} use the Lipschitz continuity property to establish that methods like SmoothGrad(gradient-based) and Continuous-LIME (C-LIME), a variant of LIME (perturbation-based method) for continuous data, are robust in expectation. The caveat here is that these methods depend on many perturbed samples for their working. 

\vspace{-6pt}\subsection{Common Techniques}
\label{subsubsec:common-tech}
Some of the common techniques to obtain attributional robustness (similar to model generalization) are: \begin{inparaenum} \item regularization of the loss function; \item data augmentation; and \item combination of regularized loss function and data augmentation \end{inparaenum}. Briefly (1) focuses on better regularization of the loss function with unperturbed data to achieve attributional robustness, (2) finds ways to augment data which allows the network to gain robustness, and (3) combines (1) and (2) which is a natural extension of the two methods.

\subsubsection{Better Loss Function Construction}
\label{para:regular-loss}

We first look into loss functions designed for better attributional robustness in detail. Table \ref{tab:loss-regular} summarizes these approaches.

Section \ref{subsec:attack-intuition} notes that one reason for models to be vulnerable to attacks is the complex boundaries learned by deep models. \citet{dombrowski2019geometry,WangWRMFD20} points out the fragility of an interpretation method and analyzes its reasons from the angle of geometry. Model regularization as a technique has been used (in addition to other techniques like auto-encoders and dropout learning approaches) to improve the non-smooth characteristics of the network and enhance the stability of interpretation. The generalizability of regularizers also helps ML algorithms maintain stable outputs. \citet{SchwartzAK20} noted that regularization would have a minimum impact on model accuracy but improves the general attribution stability leading to a more robust generalized model. Generally, a regularizer $R(\theta)$ is added to a loss function to constraint the coefficients($\theta$) to increase the sparsity (through $\ell_{1}$, $\ell_{2}$, dropout, Elastic Net etc.) as follows: $\underset{\theta}{min} \mathbb{E}_{(x,y)\sim \mathcal{D}}[\mathcal{L}(\theta, x, y) + \lambda R(\theta)]$.

\begin{figure}
\centering
\sbox{\measurebox}{%
  \begin{minipage}[b]{.4\textwidth}
  \subfloat
    []
    {\label{fig:figA}\includegraphics[width=\textwidth,height=\textwidth]{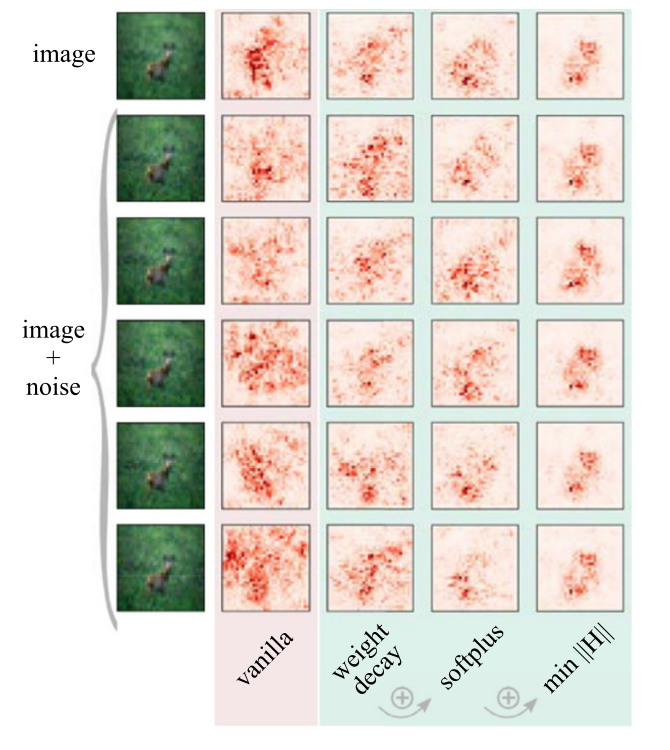}}
  \end{minipage}}
\usebox{\measurebox}\qquad
\begin{minipage}[b][\ht\measurebox][s]{.48\textwidth}
\centering
\subfloat
  []
  {\label{fig:figB}\includegraphics[width=\textwidth,height=0.38\textwidth]{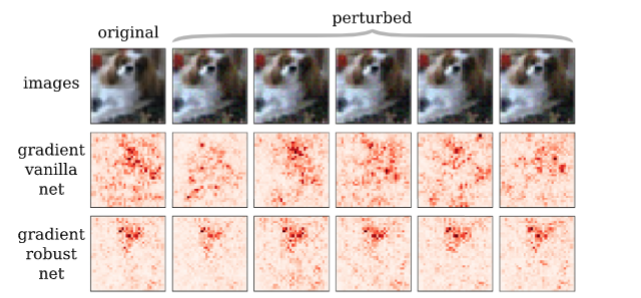}}

\vfill

\subfloat
  []
  {\label{fig:figC}\includegraphics[width=\textwidth,height=0.42\textwidth]{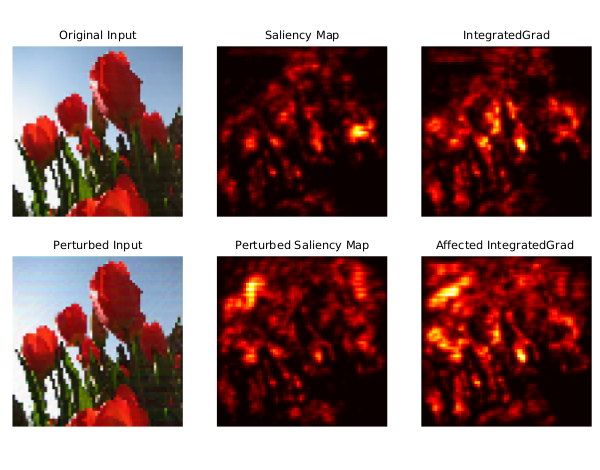}}
\end{minipage}
\caption{\protect Smooth geometry helps in robust attributions. (a) \protect \citet{dombrowski2019geometry} geometry smoothing: weight decay, smooth, (b) \protect \citet{dombrowski2019geometry} smoothing the geometry by training a network using weight decay, smoothplus activation and hessian minimization. (c) \protect \citet{WangWRMFD20} Geometry smoothing using surface smoothing regularizer. figure reproduced from respective papers.}
\label{fig:Test}
\vspace{-10pt}
\end{figure}

\citet{DombrowskiAMK22} provide a theoretical framework from which they derive insights and suggest several techniques that provide robustness to attacks \begin{inparaenum}\item weight decay \item smooth activation function \item minimize the Hessian of the network to input \end{inparaenum}reduce curvature while incorporated in the training procedure(Figure \ref{fig:Test}(a)). They reason by showing weight decay flattens the angles between piece-wise linear functions avoiding errors caused by the gradient map being too sensitive to subtle perturbations. Also, on the observation that weights of a neural network affect the Hessian of the network, the training procedure is modified such that a small value of the Frobenius norm of the Hessian is part of the objective. We notice that the explanations of the robust network are more resilient to input perturbations as in Figure \ref{fig:Test}(b). $\mathcal{L}_{0}+\zeta \sum_{x \in \mathcal{T}}||H||_{F}^{2}(x)$ where $\zeta$ is a hyperparameter regulating how strongly the Hessian norm is minimized and $\mathcal{L}$ the loss function, $x \in \mathcal{T}$ the input samples from the training set $\mathcal{T}$ and $||H||_{F}$ the Frobenius norm of the Hessian. This approach is also considered by \citet{WangWRMFD20} as a Smooth Surface Regularization (SSR) which aims to decrease the difference between saliency maps for similar inputs around a neighborhood by minimizing the Hessian matrix ${H_x}$. This improves the explanations fidelity and establishes its transferability as in Figure \ref{fig:Test}(c).
\citet{ChalasaniC00J20} used this intuition to relate robust attributional map with the property of \textit{sparsity} and \textit{stability}. Thus they formalize the loss function to consider the $\ell_{1}$-norm of the change in IG map should be small for adversarial samples to achieve the objective.



\begin{table}[h!]
    \centering
    \resizebox{\textwidth}{!}{  
    \begin{tabular}{||c|c|c||}
    \hline
    Reference  & Technique & Formulation \\
    \hline \hline
    
    \citet{dombrowski2019geometry} & Regularization & $\mathcal{L}(\theta; x+\delta,y)+\zeta \sum_{x \in \mathcal{T}}||H||_{F}^{2}(x)$ \\
    \hline
    \citet{WangWRMFD20} & Regularization & $\underset{\theta}{\min} \mathbb{E}_{(x,y)\sim \mathcal{D}}[\mathcal{L}( x, y; \theta) + \beta s \underset{i}{\max}\xi_{i}]$ \\
    \hline
    \citet{ChalasaniC00J20} & Regularization & $\mathbb{E}_{(x,y) \sim \mathcal{D}} [ \mathcal{L}(x,y;\theta) + \underset{||x+\delta-x||_{\infty} \leq \epsilon}{\max} ||IG^{\mathcal{L}_{y}}(x,x+\delta)||_{1}]$ \\
    \hline
    \citet{Chen0RLJ19} & Regularization & IG-Norm: $\label{tab:chen-rar-1} \underset{\theta}{\min} \mathbb{E}_{(x,y)\sim \mathcal{D}}[\mathcal{L}(\theta; x, y) + \lambda \underset{x' \in \mathcal{N}(x,\epsilon) \in S}{\max} s(IG^{l_{y}}_{h}(x,x+\delta;r))]$  \\
    \hline
    \citet{ivankay2020far} & Regularization & AAT: $\underset{\theta}{\min} \mathbb{E}_{(x,y)\sim \mathcal{D}} \underset{\delta \in S}{\max} \{\mathcal{L}({\theta; x}, y)+\lambda \cdot \operatorname{PCL}[\operatorname{IG}({x+\delta}, 0), \mathrm{IG}(x, 0)]\}$ \\    
    \hline
  
    \citet{MadryMSTV18} & Augmentation & $\underset{\theta}{\min} \mathbb{E}_{(x,y)\sim \mathcal{D}}[\underset{\delta \in S}{\max} \mathcal{L}(\theta; x+\delta, y)]$ \\
    \hline
    
    \citet{Chen0RLJ19} & Combined & IG-SUM-Norm: $\label{tab:chen-rar-2} \underset{\theta}{\min} \mathbb{E}_{(x,y)\sim \mathcal{D}}[\mathcal{L}(\theta, x+\delta, y) + \lambda \underset{x+\delta \in \mathcal{N}(x,\epsilon) \in S}{max} s(IG^{l_{y}}_{h}(x,x+\delta;r))]$  \\
    \hline
    \citet{ivankay2020far} & Combined & AdvAAT: $\underset{\theta}{\min} \mathbb{E}_{(x,y)\sim \mathcal{D}} \underset{\delta \in S}{\max} \{\mathcal{L}({\theta, x+\delta}, y)+\lambda \cdot \operatorname{PCL}[\operatorname{IG}(x+\delta, 0), \mathrm{IG}(x, 0)]\}$ \\
    \hline
    \citet{SarkarSGB21} & Combined & $\mathbb{E}_{(x,y) \sim \mathcal{D}} [\underset{\theta}{\min}  \mathcal{L}\left(x+\delta, \mathbf{y} ; \theta\right)+\lambda\left(\mathcal{L}_{CACR}+\mathcal{L}_{WACR}\right)]$ \\
    \hline
    & & inner maximization: $\underset{x^{\prime} \in N(X, \epsilon)}{max} \mathcal{L}\left(x+\delta, y ; \theta\right)+S(\nabla \tilde{\mathcal{A}})$ \\
    \hline
    \citet{wang2022exploiting} & Combined & $\mathbb{E}_{(x,y) \sim \mathcal{D}} [\mathcal{L}(\theta;x+\delta, y)+\lambda(1-\cos (\operatorname{IG}({x}), \operatorname{IG}({{x+\delta}}))]$ \\
    \hline 

    \end{tabular}
    }
    \caption{Attributional defenses with: (1) Regularization: Loss Function regularization, (2) Augumentation: Data Augmentation Technique and (3) Combined: Combination of both (1) and (2).}
    \label{tab:loss-regular}
    \vspace{-26pt}
\end{table}

\citet{Chen0RLJ19} in IG-Norm method added a regularizer term (Table \ref{tab:loss-regular}) to minimize the norm of differences between the attributional maps using IG of the original and perturbed image to achieve attributional robustness. \citet{ivankay2020far} also took a similar direction to modify the training objective to achieve maximal attribution correlation within a small local $L_{\infty}$ neighborhood of input by minimizing the {Pearson correlation coefficient (PCC)}. This has been discussed in more detail in Section \ref{para:regular-loss-combine}.

\begin{wrapfigure}[30]{R}{0.66\textwidth}
    \vspace{-20pt}
     \centering
     \begin{subfigure}[b]{0.58\textwidth}
         \centering
         \includegraphics[width=\textwidth]{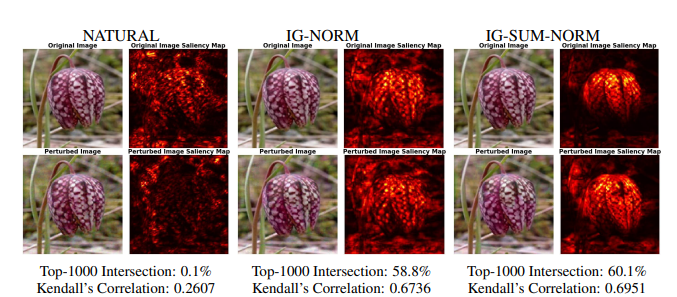}
         \label{chen:ig-norm}
     \end{subfigure}
     \vspace{-40pt}
     \begin{subfigure}[b]{0.58\textwidth}
         \centering
         \includegraphics[width=\textwidth, height=0.4\textwidth]{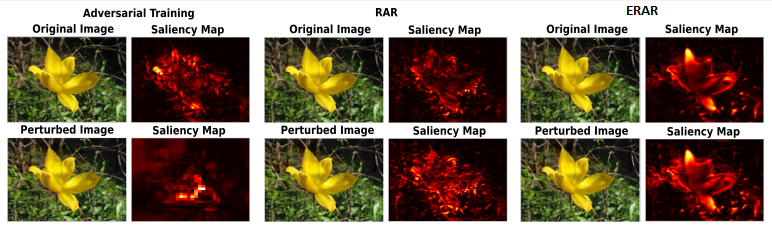}
         \label{sarkar:compare-madry}
     \end{subfigure}
     \vspace{-30pt}
     \begin{subfigure}[b]{0.58\textwidth}
         \centering
         \includegraphics[width=\textwidth, height=0.4\textwidth]{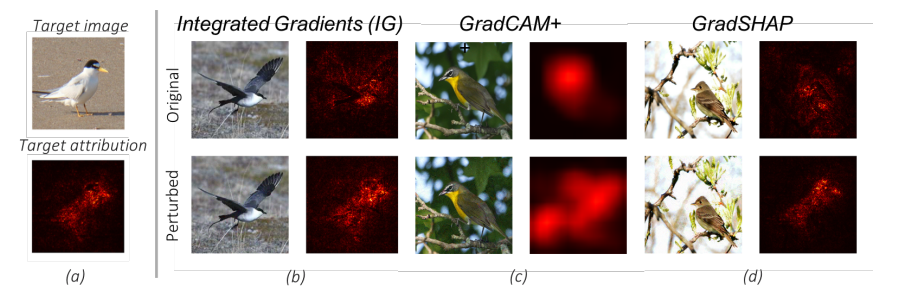}
         \label{singh:input-allign}
     \end{subfigure}
     \vspace{10pt}
        \caption{(top) \protect \citet{Chen0RLJ19} formulate multiple regularizations of the loss function. IG-Norm adds a regularization term and IG-SUM-NORM combines regularization of the loss function with data augmentation. (middle) \protect Comparison of the quality of explantions between \citet{MadryMSTV18},  \citet{Chen0RLJ19} and \citet{SarkarSB21}. (bottom) Robust attribution method improves attribution map along with ensuring robustness. \protect \citet{SinghKMSBK20}. Figures reproduced from respective papers.}
        \vspace{15pt}
        \label{fig:reg-effects-robust-attribution}
\end{wrapfigure}
    
\subsubsection{Augmentation Techniques} \label{para:data-aug}

Adversarial training is a training approach where the model is trained with adversarially perturbed samples, and the loss function is unchanged, or it is the standard $\mathcal{L}$(cross-entropy). The purpose is to make the model well-informed about adversarial samples and improve its robustness during its application. Adversarial training is given as $\underset{\theta}{min} \mathbb{E}_{(x,y)\sim \mathcal{D}}[\underset{\delta \in S}{max} \mathcal{L}(\theta, x+\delta, y)]$, which is a two-step process \begin{inparaenum} \item outer minimization; and an \item inner maximization. \end{inparaenum} The inner maximization is typically used to identify a suitable perturbation that achieves the objective of an adversarial attack, and the outer minimization seeks to minimize the expected loss. Currently, the strongest adversarial training method is using Projected Gradient Descent(PGD) attack in the inner maximization step as given in \citet{MadryMSTV18}. Section \ref{sec:adv-connect} details the connection between adversarial robustness and attributional robustness of deep learning systems.

\subsubsection{Combination of Tuned Loss Function with Data Augmentation} \label{para:regular-loss-combine}

Attempts have been made to combine regularization of the loss function and better augmentation to obtain the best of both worlds.

\citet{Chen0RLJ19} introduced IG-SUM-Norm method in addition to IG-Norm discussed in Section \ref{para:regular-loss} by combining the sum size and norm size function and adding an extra IG term to $\mathcal{L}(x^{\prime})$ as $\beta||\mathrm{IG}^{\ell_y}\left(x, x^{\prime}\right)||$. \citet{wang2022exploiting} follow a similar technique of maximizing the cosine similarity between the original and perturbed attributions. But, the regularization term added by \citet{Chen0RLJ19} suffers from the issue of vanishing second derivative for the ReLU function. This makes the optimization unstable. To address this issue \citet{SarkarSB21} used a triplet loss with softplus non-linearities. This approach focuses on pixels that attribute highly to the prediction of the true class (positive class pixels). In general, in any attribution map, there are only a few positive class pixels compared to negative class pixels. So, contrastive learning cannot improve until focus is also given to negative and positive class pixels. \citet{SarkarSB21} applied this concept by using a regularizer that forces the true class distribution to be skew-shaped and the negative class to behave uniformly. They also tried to impose a bound to change in the pixel attribution weight as indistinguishable changes are made to the input via another regularizer. Thus we observe that using appropriate loss regularization helps to obtain better attribution maps robust to attack by \citet{ghorbani2017fragile}  in Figure \ref{fig:reg-effects-robust-attribution}.
   
Based on \citet{EtmannLMS19} approach of looking at the alignment of the image and saliency map, \citet{SinghKMSBK20} think along the lines of maintaining a spatial alignment between the input image and its attribution map. They use the soft-margin triplet loss to increase the spatial correlation of input with its attribution map. 
Further, \citet{ivankay2020far} point out that \citet{ManglaSB20} compute the inner product between input and gradients, thus coupling input data and gradient domains. This approach can not be straightforwardly defined for non-continuous inputs like categorical variables, text, or other constraint inputs in multimodal problems. Also, training methodologies using joint optimization of adversarial and attributional robustness do not allow separate analyses of these notions. Thus, they try to come up with two different training methodologies one for attributional robustness and another for joint adversarial and attributional robustness. 
The training objective is modified to achieve maximal attribution correlation within a small local $L_{\infty}$ neighborhood of input by minimizing the {Pearson correlation coefficient (PCC)}. The a robust training loss (AdvAAT) can be defined as: $ \underset{\theta}{\arg \min } \sum_{x \in \mathcal{D}} \max _{\|\tilde{x}-x\|_{\infty}<\varepsilon}\{\mathcal{L}(\tilde{\theta, x}, y)+\lambda \cdot \operatorname{PCL}[\operatorname{IG}(\tilde{x}, 0), \mathrm{IG}(x, 0)]\}$ where, IG is used as attribution map and goal is to achieve robust prediction as well as target attribution correlates highly with the saliency maps of the unperturbed inputs. Here, $\operatorname{PCL} = 1 - (\operatorname{PCC}+1)/2$ is the loss derived from Pearson correlation coefficient $\operatorname{PCC}$.

\subsection{Techniques for Better Attribution Maps} \label{subsubsec:saliency-maps}
Attribution methods map features in input data to their corresponding contribution to the model’s prediction. Many attribution methods, described in earlier sections suffer from several limitations \cite{SundararajanTY17,Lu0XN21}. One such key limitation is that some explainability methods are independent of models, creating a disconnection between the model and the explanations produced. Also, as many popularly used post-hoc explainability methods are based on gradient backpropagation from output to input layer to compute feature importance, these suffer from known backpropagation-related issues such as gradient saturation. Gradient-based explanation methods suffer from the problem of importance isolation (the evaluation of the feature's importance happens in an isolated fashion), implicitly assuming that the other features are fixed. Perturbation sensitivity is also very high, as \citet{ghorbani2017fragile,KindermansHAASDEK19,kindermans2017} show that even imperceptible, random perturbations or a simple shift transformation of input data may significantly change the saliency maps. Therefore, we discuss the following methods which work towards ensuring that the generation of attribution maps is robust.

\subsubsection{Saliency Map Aggregation}
Ensemble methods reduce the variance and bias of machine learning models, which improves defense against attribution methods \cite{Rieger2020}. Based on this observation, several works \cite{Rieger2020, SmilkovTKVW17, Si21, Lu0XN21, Liu2021cert,manupriyaMJB22}  proposed different ways to aggregate explanation methods to make networks robust against attacks. \citet{Rieger2020} propose AGG-Mean method, which does a simple averaging over all explainable methods using normalized inputs. $\mathcal{I}_{(j,n)}$ is the explanation obtained for $x_{n}$ with explainer method $E_{j}$ and the mean aggregate explanation,  $\bar{E}_n=\frac{1}{J} \sum_{j=1}^J E_{(j, n)}$. They hypothesize the non-transferability of attacks across explanation methods, by observing that this way of aggregation can better preserve the attribution map under an attack. They also theorize that averaging the diverse set of explanation methods creates similar smoothness behavior as that of SmoothGrad \cite{SmilkovTKVW17}.

\citet{Si21} propose a method to generate a robust attribution map using a preprocessing technique, which does not change the training or attribution methodology. The attribution map is obtained by adding a gaussian noise to the input space and multiplying the averaged gradient and feature maps obtained by passing it through the deep neural network model. They hypothesize that averaging many times will restore the perturbed input. \citet{Lu0XN21} similarly add some noise to the input samples to produce perturbed samples, which follow the same distribution as the input distribution. Then, the saliency maps for the perturbed samples are obtained and aggregated which seems to capture inter-feature dependence and is robust against adversarial perturbations.
\citet{Liu2021cert} use a similar approach of adding noise to input images and aggregating the output saliency maps of the neural network model using a voting rule. They finally prove its robustness using the notion of \'{R}enyi differential privacy  which is based on \'{R}enyi divergence that measures difference between distributions. \citet{manupriyaMJB22} propose a submodular strategy to ensemble attribution maps which also provides more robust and sparse explanations.

All the above methods provide theoretical and experimental evidence, supporting generating stable explanation maps to be an effective way to ensure robustness. But, these methods have limitations. Some of these methods make an assumption about prior knowledge like the attack type \cite{Liu2021cert} to be known and hence, may damage the original interpretation effect to some extent in the face of unperturbed input.

\subsubsection{Establishing a Connection Between Model and Attribution Method}
Though model-agnostic methods (e.g. perturbation-based methods such as LIME) have the advantages of flexibility, they do not use the internal structure of the model. In contrast, model-specific methods like GradCAM, IG, and SmoothGrad make use of the model's structure. So, to counter adversarial examples, the primary focus is on using the model's structure effectively to preserve the actual saliency map. 
\begin{wrapfigure}[9]{r}{0.5\textwidth}
    \vspace{-20pt}
    \centering
    \includegraphics[width=0.3\textwidth]{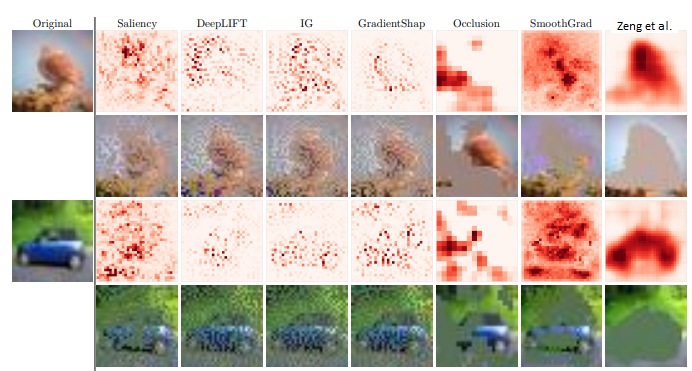}
    \vspace{-12pt}
    \caption{A comparison of attribution maps obtained from different methods, and even rows show images when the top 50\% fraction of pixels deemed most important by each attribution method is replaced with the mean. It observes that better alignment between the model and attribution map produces better attribution maps. Reproduced from \citet{ZengKE021}}
    \label{fig:posthoc-model-align}
\end{wrapfigure}
Several works \cite{Lu0XN21,ZengKE021,HuaiLMYZ22} have tried approaches to establish a connection between the model and the interpretability method. 
\citet{ZengKE021} established the connection between the attribution map and the DNN model. They first obtain an adversarially robust model (using \citet{MadryMSTV18} method) $\mathcal{F}'$ with the same architecture as that of original model $\mathcal{F}$. They use $\mathcal{F}'$ to align the attributions with $\mathcal{F}$ by minimizing the $\ell_{2}$ distance between them. The explainer model is trained using the following objective $\min _\theta \mathbb{E}_{(x, y) \sim \mathcal{D}}\left[\mathcal{L}(\theta, x, y)+\lambda\left\|\mathcal{I}(\mathcal{F},x)-\mathcal{I}\left(\mathcal{F}', x\right)\right\|_2\right]$ where $\lambda$ is the loss weight for balancing and $\mathcal{L}(\theta, x, y)$ is the cross-entropy loss. By aligning $\mathcal{F}'$ with $\mathcal{F}$, the attribution stability improves. See Figure \ref{fig:posthoc-model-align} for an example.

Table \ref{table:attack-metrics} summarizes evaluation metrics, data sets and approaches for different methods to ensure robustness. 
Attributional robustness, a less-studied area but highly relevant for real-world industrial deployments, has attracted more profound interest in recent years from the industry, academia, and regulators alike. We discuss the issues and future direction of current robustness methods in Section \ref{sec:conclusion}.

\newcolumntype{P}[1]{>{\RaggedRight\arraybackslash}p{#1}}

\begin{table}[htbp]
\centering
\resizebox{0.8\textwidth}{!}{
\begin{tabular}{||P{3cm} | P{2.3cm} | P{5cm} | P{3.5cm}||} 
 \hline
 Reference & Method to Ensure Robustness & Evaluation Metric & Dataset \\ 
 \hline\hline

    \citet{SinghKMSBK20} & CAR & Top-$k$ Intersection, Kendall’s $\tau$ & CIFAR-10, SVHN, GTSRB, Flower \\ 
 \hline
 

    \citet{dombrowski2019geometry} & CAR & SSIM, PCC, MSE & ImageNet, CIFAR-10\\ 
 \hline
 
    
    \citet{HuaiLMYZ22} & CAR, CMA & Top-$k$ Intersection & MNIST, CIFAR-10, AT\&T \\ 
 \hline
 

   \citet{AgarwalJAU0L21} & LC & Top-$k$ Intersection & Bankruptcy, Online Shopping \\ 
 \hline 
 

   \citet{WangWRMFD20} & LC, LFC & Spearman's $\rho$ , Top-$k$ Intersection, Mass Center Dislocation, Cosine Distance & Flower, CIFAR-10 \\ 
 \hline
 

   \citet{Chen0RLJ19} & LFC, CLA & Top-$k$ Intersection, Kendall’s $\tau$ & MNIST, Fashion-MNIST, GTSRB, Flower \\ 
 \hline
 

    \citet{SchwartzAK20} & LFC & Pixel-wise Sum of Squared Distances & CIFAR \\ 
 \hline
 

   \citet{DombrowskiAMK22} & LFC & SSIM, PCC, MSE & CIFAR-10, ImageNet \\ 
 \hline
 

   \citet{ChalasaniC00J20} & LFC & Gini Index & MNIST, Fashion-MNIST, CIFAR-10, Mushroom, Spambase \\ 
 \hline
 

   \citet{SarkarSB21} & CLA & Top-$k$ Intersection, Kendall’s $\tau$, Spearman's $\rho$ & MNIST, Fashion-MNIST, GTSRB, Flower \\ 
 \hline
 

    \citet{ivankay2020far} & CLA & Top-$k$ Intersection, Kendall’s $\tau$ & MNIST, Fashion-MNIST, CIFAR-10, GTSRB, Restricted ImageNet \\ 
  \hline
  
  \citet{wang2022exploiting} & CLA & Top-$k$ Intersection, Kendall’s $\tau$ &  MNIST, Fashion-MNIST, CIFAR-10\\ 
  \hline
  

    \citet{Liu2021cert} & SMA & Overlapping Ratio between the Top-$k$ Components & PASCAL Dataset \\
 \hline
 
 
  \citet{SmilkovTKVW17} & SMA & Energy Ratio & ILSVRC 2012 Dataset \\
 \hline
 
 
  \citet{Rieger2020} & SMA & MSE, PCC, Top-$k$ Intersection & ImageNet \\
 \hline
 

 \citet{Lu0XN21} & CMA & Fidelity Metric & ImageNet, Stanford Sentiment Treebank \\
 \hline
 

 \citet{ZengKE021} & CLA, CMA & ROAR, Sanity Check & CIFAR-10, TinyImageNet \\
 \hline
 
\end{tabular}
}
\caption{Summary of different approaches used to ensure attributional robustness, along with the metric for their evaluation and the datasets on which the evaluations were conducted. The notations used in the table represent the following methods respectively. CAR: Certified Attributional Robustness, LC: Lipschitz Continuity, LFC: Better Loss Function Construction, CLA: Combination of Tuned Loss Function with Data Augmentation, SMA: Saliency Map Aggregation, CMA: Establishing a Connection between Model and Attribution Method}


\label{table:attack-metrics}
\vspace{-8mm}
\end{table}

\vspace{-5pt}\section{Connecting Explanations to Adversarial Robustness} \label{sec:adv-connect}

\begin{figure}[ht!]
  \centering
  \includegraphics[width=\linewidth]{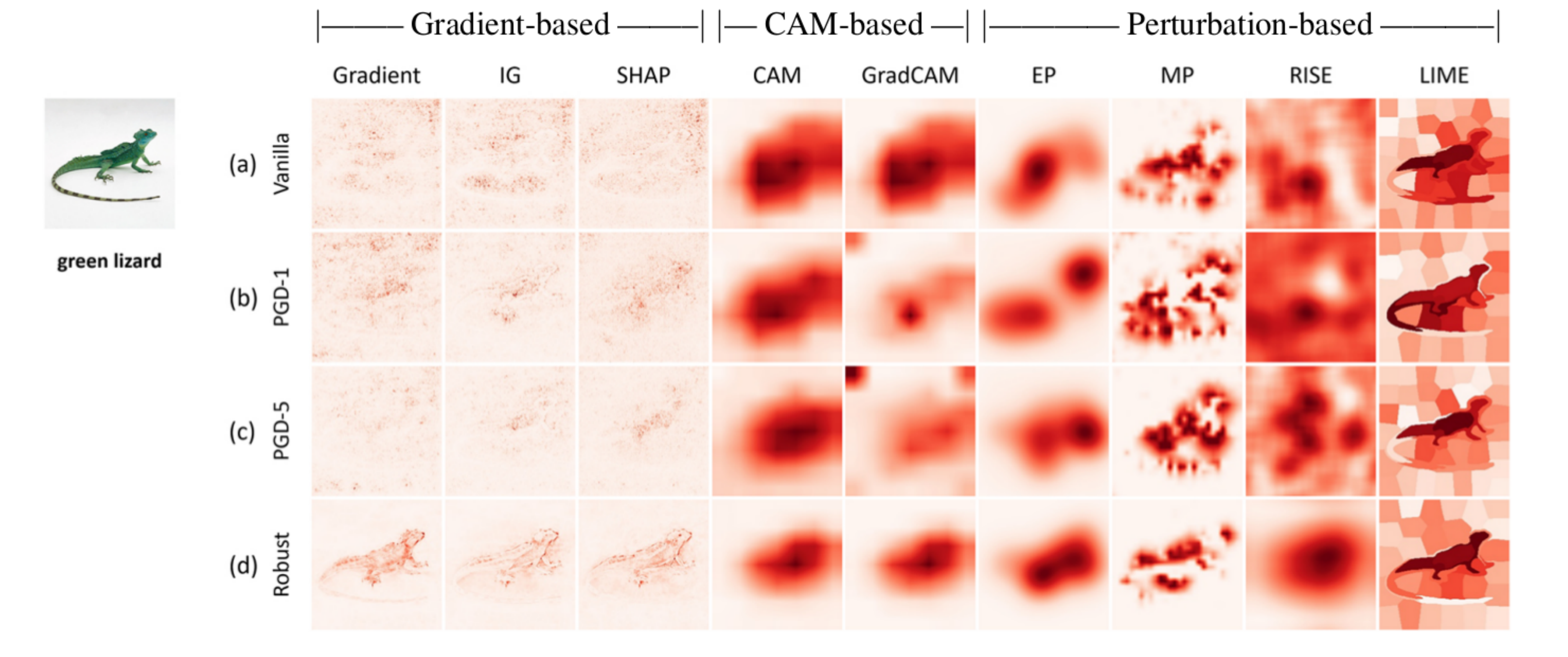}
  \vspace{-7pt}
  \caption{Detailed comparison of attribution maps obtained from 9 different explanation methods for four differently trained ResNet-50 models on ImageNet data for the same input image and target label of ``green lizard". Read from top to bottom row-wise: (a) naturally trained model; (b–c) same architecture but adversarially trained with PGD of different step sizes like PGD-1, PGD-5 (i.e. PGD with step sizes 1 and 5, respectively); and (d) adversarially trained with strongest PGD version. Note that as the network is trained adversarially with a PGD attack of gradually increasing strength, the gradient-based methods tend to become less noisy and more interpretable. Reproduced as given in \protect \citet{Nourelahi2022} [Fig 1].}
  \label{sample-connect-adv}
  \vspace{-5mm}
\end{figure}

The key concerns related to attributions that we discussed in earlier sections are that: \begin{inparaenum} \item the attribution maps have to be of good quality, and \item they should be robust under perturbations\end{inparaenum}. It is observed that adversarially robust models obtain the best explanation maps under different attribution methods (Figure \ref{sample-connect-adv} from \citet{Nourelahi2022}) as they seem to be more focused on the objects related to the input's ground-truth class. This section explores the interplay between explanations and adversarial robustness of neural network models, primarily from an image input perspective where adversarial robustness is more studied. 
\begin{figure}[!ht]
     \centering
     \begin{subfigure}[b]{1.0\textwidth}
         \centering
         \includegraphics[width=\textwidth]{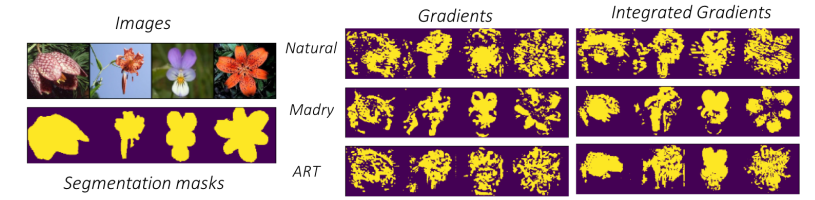}
         \label{sarkar:segmentation}
     \end{subfigure}
    \newline
     \begin{subfigure}[b]{1.0\textwidth}
         \centering
         \includegraphics[width=\textwidth]{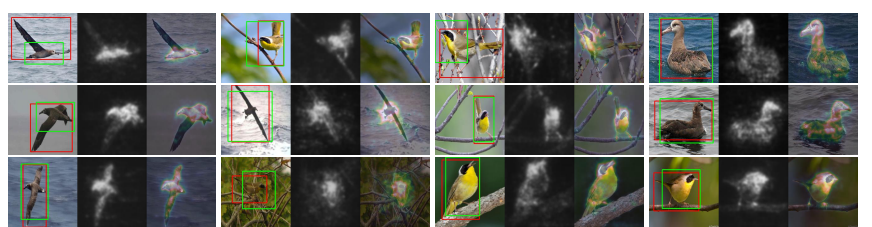}
         \label{singh:object-detect}
     \end{subfigure}
        \caption{Effects of robust attributions on other tasks. (top) \protect \citet{SarkarSB21} Good attribution helps in good segmentation mask. (bottom) Object detection. \protect \citet{SinghKMSBK20} Good attribution can aid in unsupervised object detection task. Figures reproduced from respective papers.}
        \label{fig:tasks-effects-robust-attribution}
        \vspace{-5mm}
\end{figure}

\begin{figure}[ht!]
  \centering
  \begin{subfigure}[b]{0.60\textwidth}
  \centering
        \includegraphics[width=0.9\textwidth, height=0.6\textwidth]{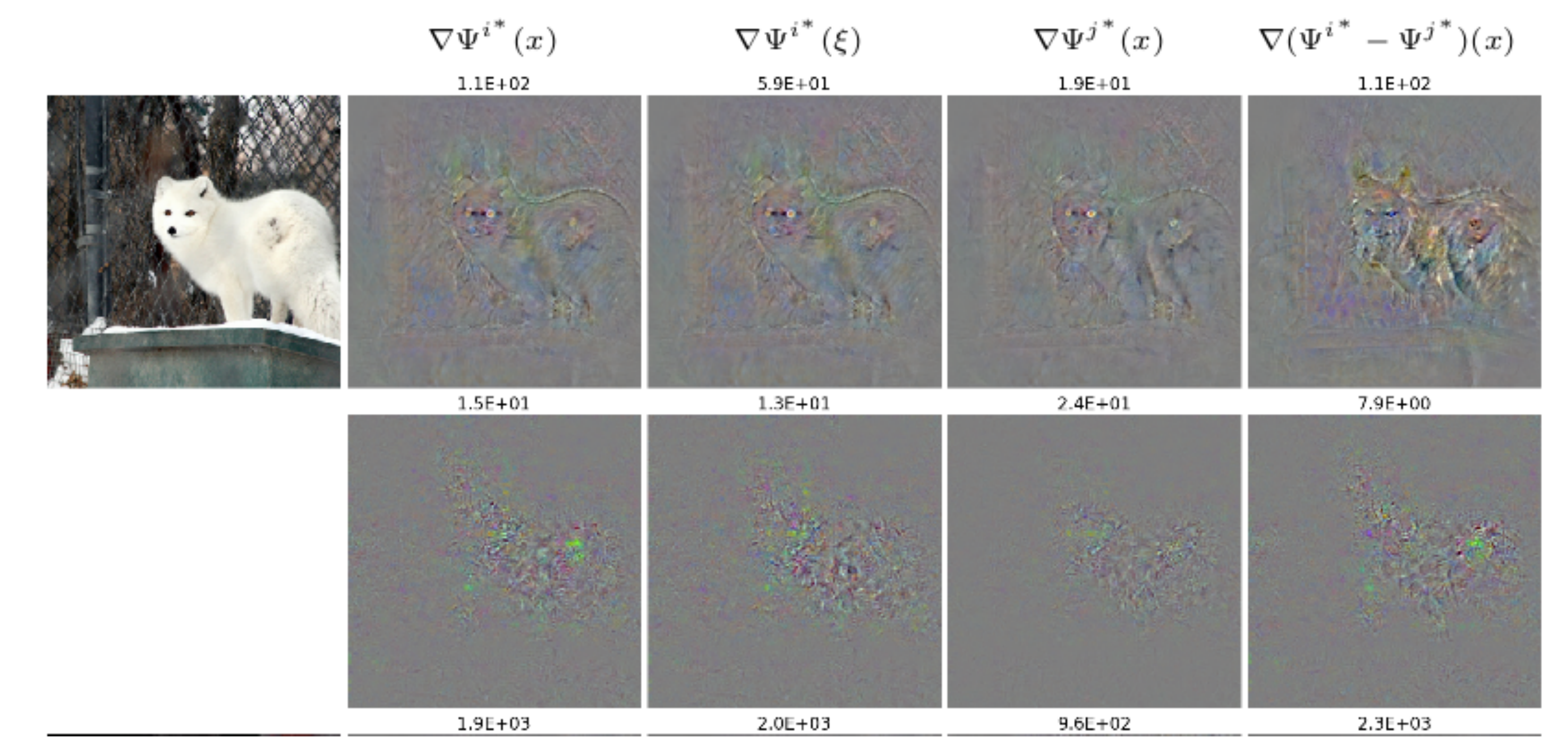}
  \end{subfigure}
  \begin{subfigure}[b]{0.39\textwidth}
  \centering
        \includegraphics[width=1.0\textwidth, height=0.9\textwidth]{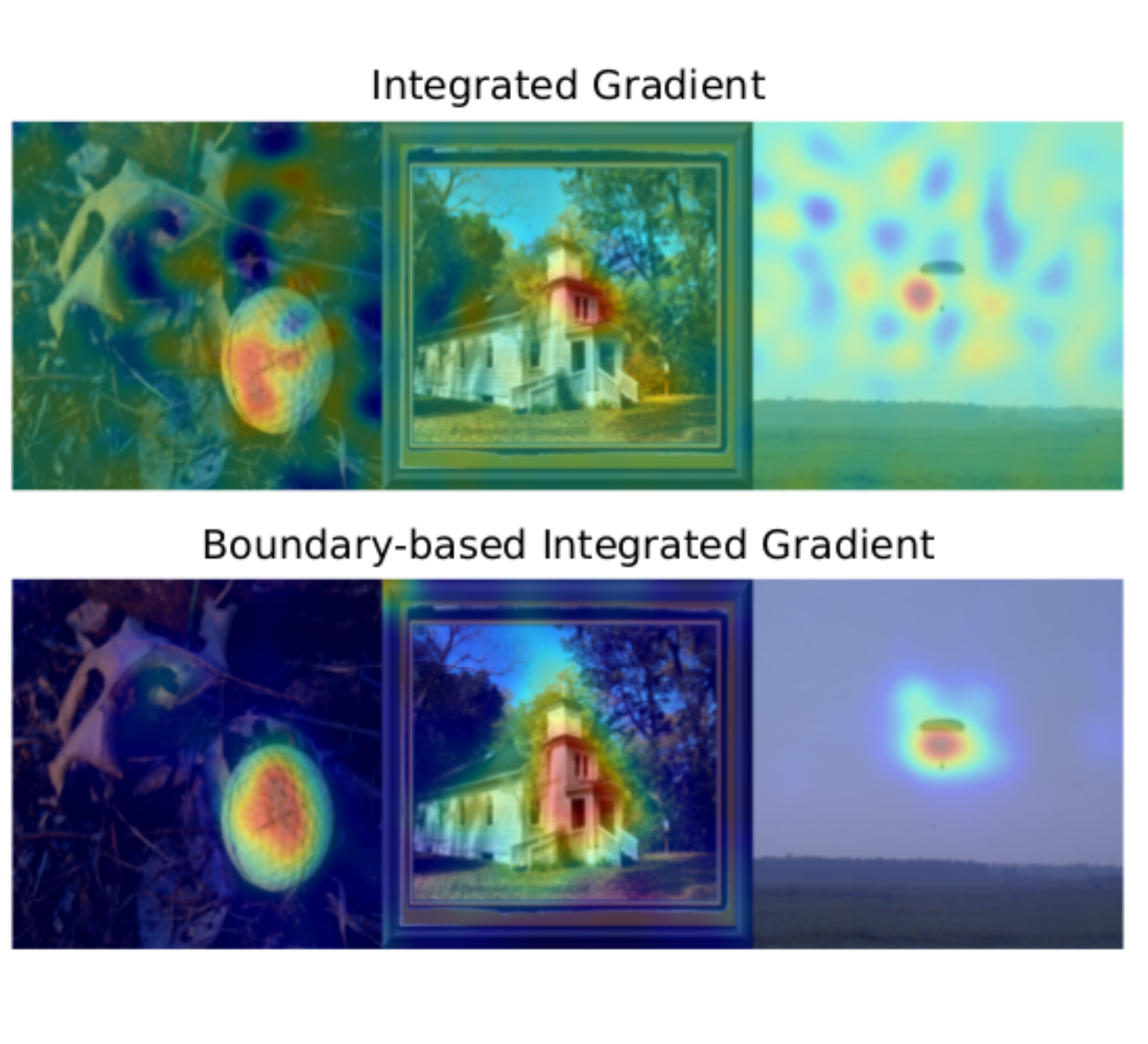}
  \end{subfigure}
  \caption{\textit{(Left)} \protect \citet{EtmannLMS19} observed that the alignment between the gradient/saliency map and the image is a good indicator of the adversarial robustness of the model; \textit{(Right)} \protect \citet{WangWRMFD20} utilize the boundary alignment property to obtain an enhanced IG -- Boundary based IG (BIG) -- which provides a sharper, more focused and less noisy saliency map. Figures reproduced from respective papers.}
  \label{sample-grad-align}
  \vspace{-5mm}
\end{figure}

We first look at various aspects satisfied by an explanation map obtained from an adversarially robust model and
understand the multiple connections between the properties of explanations and those satisfied by an adversarially robust model. While most studies are based on image classification, \cite{SarkarSB21, ManglaSB20} show that their approach also addresses the robustness of attributions for other tasks like segmentation and object detection as illustrated in Figure \ref{fig:tasks-effects-robust-attribution}.

\citet{ChalasaniC00J20} found that attribution maps obtained from a robust adversarial model satisfied two critical properties: \textit{sparsity} and \textit{stability}. Sparseness ensures attributions of irrelevant or weakly relevant features should be negligible, thus resulting in concise explanations concentrated on the significant features. Stability ensures the map should not vary significantly within a small local neighborhood of the input. They capture this property by ensuring the worst-case $\ell_{1}$-norm of the change in IG will be small: $\underset{x' \in \mathcal{N}(x,\epsilon)}{max} ||IG^{\mathcal{F}}(x',u)-IG^{\mathcal{F}}(x,u)||_{1}$, where $\mathcal{N}(x,\epsilon)$ denotes a suitable $\epsilon$-neighborhood of x, and where $u$ is an appropriate baseline input vector.


\subsection{Why Do Adversarially Trained Models Aid Better Explanations?} \label{subsec:adv-training}


A few existing works provide reasoning for why an adversarially trained model aids better explanations focus on the ``alignment" of an image and its attribution map \cite{EtmannLMS19,WangWRMFD20,ChalasaniC00J20}. The work by \citet{EtmannLMS19} is the first step in that direction. 

\noindent \paragraph{Alignment Between Image and Attribution Map} \label{subsec:adv-training:align}
\citet{EtmannLMS19} find that the saliency map is more aligned with the input image $x$ for a robust model. They consider a linear binary classifier $\mathcal{F}: X \rightarrow \{-1,1\}$, where $\mathcal{F}(x)=sgn(\Psi(x))$, where $\Psi: X \rightarrow \mathbb{R}$ is differentiable in $X$. They call $\nabla \Psi$ as the saliency map of $\mathcal{F}$ concerning $\Psi$ in x and is given as $\frac{|\langle x , \nabla \Psi (x) \rangle|}{||\nabla \Psi(x)||}$. They observe that the model's increased robustness can be correlated with the high alignment of the image with the saliency map (refer to Figure \ref{sample-grad-align}left for an example). The linear approximation of the prediction near a sample is given by $\mathcal{F}(x_{i} + \epsilon) \sim \mathcal{F}(x_{i}) + \epsilon^{T} \nabla_{x_{i}} \mathcal{F}(x_{i})$.



\noindent \paragraph{Adversarially Robust Models have Smooth Decision Boundary} \label{subsec:adv-training:smooth}
\citet{WangWRMFD20} find that a smooth decision boundary plays an essential role in obtaining better interpretable maps as the model's input gradients around data points align more closely with the boundaries' normal vectors when they are smooth. Thus, because robust models have softer boundaries, the results of gradient-based attribution methods, like Integrated Gradients(IG) and DeepLIFT, will capture more accurate information about nearby decision boundaries. They propose Smooth Surface Regularization (SSR) to minimize the difference between Saliency Maps for nearby points; $\underset{\theta}{min} \mathbb{E}_{(x,y)\sim \mathcal{D}}[\mathcal{L}(\theta, x, y) + \beta s \underset{i}{max}\xi_{i}]$, where $\underset{i}{max}\xi_{i}$ is the largest eigenvalue of the Hessian matrix $\tilde{H}_{x}$ of the regular training loss $\mathcal{L}$ w.r.t to the input. They show that SSR has comparable results to an adversarially trained model and \citet{Chen0RLJ19} trained models. They also use Lipschitz continuity (explained in detail in Section \ref{subsubsec:liptschitz-conti}) to capture the smoothness of the decision boundary  achieved by differently trained networks, suggesting that adversarially trained models naturally satisfy the Lipschitz property.

\noindent \paragraph{Connection Between IG and Adversarial Training} \label{subsec:adv-training:concise}
\citet{ChalasaniC00J20} observe the connection between IG maps and adversarial training. They use the implicit definition of IG, which ensures each feature's attribution is essentially proportional to the feature's fractional contribution to the logit-change for x. Hence, if the weight-vector w is sparse for such models, then the IG vector will also correspondingly be sparse. Stable-IG Empirical Risk is formulated as $\mathbb{E}_{(x,y) \sim \mathcal{D}} [ \mathcal{L}(x,y;w) + \underset{||x'-x||_{\infty} \leq \epsilon}{max} ||IG^{\mathcal{L}_{y}}(x,x')||_{1}]$, which is equivalent to minimizing the expected $\ell_{\infty}(\epsilon)$-adversarial loss. In other words, for this class of loss functions, natural model training while encouraging IG stability is equivalent to $\ell_{\infty}(\epsilon)$-adversarial training. 
Robust models admit smoother gradient images (Figure \ref{sample-connect-adv}) and smoother activation maps, i.e., internally filtering out high-frequency input noise, which highly improves the quality of attribution maps. Robust models outperform vanilla ones in gradient-based methods but not in CAM-based or perturbation-based methods.

\subsection{Exploiting Connection between Adversarial Robustness and Attributions for Better Training Methods} \label{subsec:adv-training:exploit}
In this section, we explore the impact of the connections discussed in the previous section to obtain better adversarial training or attribution methods.

\noindent \paragraph{Using Attributions to Aid Adversarial Robustness}  
\citet{SarkarSGB21, ManglaSB20} exploited the properties of attribution maps to provide better training for better adversarial robustness than current standard methods like PGD training \cite{MadryMSTV18} and early stopping \cite{RiceWK20}. \citet{ManglaSB20} proposed Saliency-based Adversarial Training (SAT), which uses annotations such as bounding boxes and segmentation masks to be exploited as weak explanations to improve the model's robustness with no additional computations required to compute the perturbations themselves. \citet{li2022consensus} observed that the progressive saliency information regarding the input semantic features the model relies on to make a prediction is not utilized. They suggest incorporating this information into the training method to enhance robustness. They propose to optimize the discrimination of intermediate gradient-based saliency and maintain its consensus in training, which encourages the model to behave according to task-relevant features from the salient region, such as object edges in the image. The proposed Adversarially Gradient-based Saliency Consensus Training method, dubbed Adv-GSCT is $\underset{z(x)}{min} \mathcal{D}_{KL} (\mathcal{F} (x \odot z(x)), \mathcal{F}(x))$, and $\underset{z(x)}{min} \mathcal{D}_{KL} (\mathcal{F} (x \odot z(x)), \mathcal{F}(x)) + ||z(x)||_{1}$ where $\mathcal{D}_{KL}$ is Kullback-Leibler divergence measuring how a probability distribution differs from the other and, $z(x)$ represents the normalized saliency map, such that $\mathcal{F}(x') = \mathcal{F}(x)$, with $x' = x \odot z(x)$. 

\noindent \paragraph{Better Attribution Methods}
\citet{WangFD22} exploited the boundary connection discussed in the previous section to provide a better explanation method: Boundary-based Integrated Gradient (BIG), an approach that explores a linear path from the boundary point to the target point traversing different activation polytopes, $P_{i}$. The aggregation is done for each gradient $w_{i}$ along the path and weights it by the length of the path segment intersecting with $P_{i}$. Given model $\mathcal{F}$, Integrated Gradient $gIG$ and an input $x$, they define Boundary-based Integrated Gradient BS(x) as follows: $BIG(x) := gIG(x; x_{0})$, where $x$ is the nearest adversarial example to $x$, i.e., $c=\mathcal{F}(x) \neq \mathcal{F}(x')$ and $\forall x_{m} ||x_{m} - x|| < ||x_{0} - x|| \rightarrow \mathcal{F}(x) = \mathcal{F}(x_{m})$. Thus, it tries to explore the local geometry around $x$, compared to IG, which explores the model's global geometry by aggregating all boundaries from a fixed reference point.

\noindent \paragraph{Defense Based on Interpretability-Awareness} 
\citet{Boopathy0ZLCCD20} developed an ``interpretability-aware defensive scheme'' built only on promoting robust interpretation (without the need for resorting to adversarial loss minimization). They show that their defense achieves both robust classification and robust interpretation, outperforming state-of-the-art adversarial training methods against attacks of significant perturbation. Their formulation is $\underset{\theta}{minimize} \mathbb{E}_{(x,y) \sim \mathcal{D_{train}}}[\mathcal{F}_{trin}(\theta,x,y) + \gamma \tilde{D}_{worst}(x,x')]$.

\noindent \paragraph{Natural Training Made Better with Loss Function for Stable Attribution} 
\citet{ChalasaniC00J20} provided a Stable-IG Empirical Risk which is equivalent to minimizing the expected $\ell_{\infty}(\epsilon)$-adversarial loss. In other words, for this class of loss functions, natural model training while encouraging IG stability is shown to be equivalent to $\ell_{\infty}(\epsilon)$-adversarial training. This raises an exciting question about the connection between \citet{Chen0RLJ19}, who aim to train the network for attributional robustness, and \citet{MadryMSTV18}, who aim to train the network for adversarial robustness.

\vspace{-6pt}\subsection{Other Work Relating Adversarial Robustness and Attributions} \label{subsec:connect:adv-train-attr-robust}
Among other perspectives that connect adversarial robustness and explanations, \citet{ManglaSB20} studied the question if one can construct an adversarial attack from an explanation. In particular, the work generates adversarial perturbations from a given saliency map to improve robustness while training a neural network. The intuition is that the direction of adversarial perturbation can be obtained using a saliency map. The negative of the saliency map corresponding to the ground truth class, i.e. $- \nabla_{x} \mathcal{F}(x_{i},y_{i})$, can be used as a direction to perturb the input. Their experiments show that examples generated by perturbing an input in the direction of negative saliency decrease the model's accuracy. This shows that though it may be a weak adversarial direction, it has misclassification potential, and thus can be useful for adversarial attacks or defenses. 

\citet{bogun2021ensemble} presented an interesting perspective to obtaining better explainability using an ensemble of deep learning models. This ensemble is trained using a regularizer that tries to prevent an adversarial attacker from targeting all ensemble members at once by introducing an additional term in the learning objective. In addition to providing improved adversarial robustness under white-box and black-box attacks, this work shows that adversarially training a DNN ensemble can also lead to robust explanations.

\vspace{-5pt}\section{Discussions, Take-Aways and Conclusion}
\label{sec:conclusion}
In this work, we presented a comprehensive survey of methods, metrics and, perspectives to achieve robust explanations of DNN model predictions. As we showed in Section \ref{sec:intro} (see Figure \ref{fig_intro}), ensuring the stability and robustness of explanations to minor changes in input data is essential to the deployment-readiness of deep neural network (DNN) models, especially in risk-sensitive and safety-critical applications such as healthcare, autonomous navigation, security and, finance. Methods that study and ensure the robustness of explanations of DNN model predictions form a key component of the community's work towards responsible and trustworthy use of these models. While explainability methods themselves have been summarized through many literature surveys in the recent past, there has been no effort to consolidate existing work on the robustness of explanations in DNN models. This work is the first such in this direction. Many methods have been proposed to measure, attack, and defend the robustness/stability of explanations in recent years, making this a field that is not nascent or embryonic in its development. However, it is not extensively researched either, making this an opportune moment for such a survey to take note of the field as a whole, understand its current status and development, and use that knowledge to provide collated information to increase overall awareness and motivate further research.

This survey presented a detailed listing of evaluation metrics for explanation methods, with a focus on metrics used to study their robustness/stability. Among existing studies of explanation methods for DNN models, while there have been a few efforts on studying their fidelity, there has been very little concerned effort to study their robustness, thus making this work valuable. We then present an overview of various methods that have been proposed to attack and defend explanations in DNN models on different data types, including images, text, and tabular data. Considering the significant efforts over the last few years to study adversarial robustness, we also discuss the connections of adversarial robustness to explanations in general, to make this work more complete. We summarize our take-aways across this work, along with pointers to future research directions, below:
\begin{itemize}[leftmargin=*]
    \item Beginning with sound and robust explanation methods themselves, while different methods perform well on different metrics, an overall observation across the literature reviewed in this work suggests that methods incorporating smoothing or SmoothGrad seem to perform better in general. This is an important take-away from this survey, especially considering how simple it is to implement SmoothGrad, and how it can be easily integrated into existing explanation methods. This observation is intuitively understandable however, considering such an approach smoothens the explanation over a few input perturbations, thereby imbuing the model with robustness of explanations. A deeper theoretical understanding of this observation would be an interesting direction of future work.
    \item Many metrics for attributional robustness have been used including \textit{$L_2$ norm}, \textit{Spearman's rank order correlation}, \textit{top-$k$ intersection}, \textit{Kendall's $\tau$}, \textit{ROAR}, \textit{Energy Ratio}, and the \textit{Pixel-wise Sum of Squared Distances}. Among these, the most widely used are \textit{Spearman's rank order correlation}, \textit{Kendall's $\tau$} and \textit{top-$k$ Intersection}. The popularity of these metrics can be attributed to the fact that they concern themselves with the correlation between the feature rankings, and not the actual attribution values themselves. In other words, these metrics compare the set of most relevant features from either map and evaluate the intersection or correlation between these sets. Noting that the relative ranking of attribute importance is more important than the attribution value itself may be an important consideration for the design of attribution methods in the future.
    \item Metrics for attribution robustness are still in a nascent stage with few metrics like top-$k$ intersection, Spearman's $\rho$ and Kendall's $\tau$ correlations most used at this time. While these have provided the community with a good basis to start with, there is plenty of scope to expand the field with better metrics/pseudo-metrics for this purpose. For example, in image data (where these metrics have been most studied so far), each of these metrics is highly sensitive to minor changes in spatial coordinates, viz. if the attribution map changed by even one pixel with an input perturbation, these metrics could count this as a non-robust attribution. Having locally smoothed metrics for such contexts may help provide a more realistic understanding of robustness. There is no work hitherto in this respect. Moreover, metrics that are certifiable or act as certificates by themselves may be a way to go forward for ensuring high measurable standards of reliability in any field of use. 
    \item More generally speaking, Section \ref{sec:eval-explain} presents a reasonably comprehensive list of metrics that are used for evaluating explanations in DNN models. The subjective nature of explanations requires such a list of metrics, but this long list presents challenges to a user in understanding what kind of metric one ought to use in a given setting. It may benefit practitioners in the field if the process of providing explanations to model predictions is streamlined through a user interface, where the application requirements are specified. Appropriate metrics can then be suggested to a developer/end-user based on the provided requirement specifications, thus mitigating the burden of choosing metrics for studying explanations in a given application context. 
    \item The current benchmark for studying attribution attacks is the method proposed by \citet{ghorbani2017fragile}. Given this unanimous choice, a more detailed study on evaluation metrics and training approaches for attributional robustness may be required to understand the limitations of this attack and propose newer attacks -- which in turn may promote robust explanations. As stated in Section \ref{sec:attack-explain}, every attack proposed so far can be viewed as a path taken in a decision tree, with decisions being various techniques (e.g. COM shift, top-$k$), multiple spaces (e.g. input, feature), different explanation methods (e.g. IG, LIME), and input types (e.g. Image, NLP). There are yet many paths that have not been discovered and experimented with. Exploring these will only provide more ways of also making explanations robust.
    \item Section \ref{sec:ensure-robust} presented different methods proposed to obtain robust explanations with DNN models. However, the generalizability of existing defense techniques to: (i) other models (e.g.recurrent networks, transformers, attention-based frameworks, etc); (ii)  domains such as text (NLP), tabular and time-series-based datasets; (iii) unsupervised or multi-modal learning settings with high-dimensional and noisy datasets remains unstudied and maybe a critical need for the immediate future. Besides, it may be useful to have auxiliary techniques to determine what compromises the robustness of a given model; for e.g., is a given model explanation's robustness affected by input data, network, or the attribution method? Knowing this would help enhance these methods going forward.
    \item Section \ref{sec:adv-connect} studied the connection of adversarial robustness and adversarial training to explanations and their stability. In particular, this section addressed the question: How much does adversarial training aid the quality of attributions? A natural follow-up question to this discussion is: Can adversarial training guarantee attributional robustness for free? Does attributional robustness help towards adversarial robustness? This is an important question since adversarial training, which is the de facto approach to make a DNN robust to adversarial perturbations, is an expensive strategy. The use of explanations as a cheap alternative to obtain similar robustness would hence be of immense value.
    The consequent study of relationships between multiple perspectives of robustness (e.g. adversarial, attributional, spatial) also remains an unexplored problem and a potential important research direction for the community.
    \item Going beyond attributional attacks, we presented the notion of trust attacks in Section \ref{sec:attack-explain}. As stated therein, an attributionally robust model might not be trust-wise robust and vice-versa. Such trust attacks provide a new dimension to attacks on explanations and help raise concerns about the validity of explanations and how well they are in line with the predictions made. While it may be challenging for a model to be trained to address multiple vulnerabilities, testing for multiple parameters (robust explanation and robust prediction, in this case) can help identify potential biases in the model. An important future direction in this regard is to view testing of machine learning models as a separate concrete paradigm (similar to software testing) that considers all such perspectives.
    \item Existing methods for attributional attacks and defenses are based on post-hoc explanation methods. However, there has been an increasing demand for ante-hoc (intrinsically interpretable) explainable models in the community \cite{rudin2019stopexplaining,rudin2021interpretableml}. Studying the robustness of explanations in such ante-hoc interpretable models remains an open problem at this time. Building inherently explainable models that are robust, versatile, efficient, accurate, and as scalable as DNN models could be a vision to achieve for the community.
    \item Finally, the lack of robustness of explanations in DNN models continue to raise questions on the opaqueness of a neural network's decision process and the lack of a deeper understanding of how they actually work. Having stronger theoretical foundations may have a direct implication on obtaining robust models by itself.
\end{itemize}

While the research efforts over the last few years have studied evolving robustness strategies, they have also exposed several weakness around generalizability, algorithmic efficiency, and stability of explanations in DNN models. We hope this survey provides useful insights to the research community to take insights from these existing works and work towards robust explanations and trustworthy models.

\bibliographystyle{ACM-Reference-Format}
\bibliography{arxiv}
\end{document}